%% file: arxiv.tex
\documentclass[10pt,letterpaper]{article}
\usepackage[T1]{fontenc}
\usepackage[utf8]{inputenc}
\usepackage{times}
\usepackage[letterpaper,textwidth=5.5in,textheight=9in,centering]{geometry}
\usepackage[authoryear,round]{natbib}
\setcitestyle{citesep={;},aysep={,},yysep={;}}

\input{math_commands.tex}

\usepackage{hyperref}
\usepackage{url}
\usepackage{booktabs}
\usepackage{multirow}
\usepackage{graphicx}
\usepackage{subcaption}
\usepackage{afterpage}
\usepackage{capt-of}

\usepackage[most]{tcolorbox}
\usepackage{enumitem}
\usepackage{amsmath}

\usepackage{tikz}
\usetikzlibrary{positioning, arrows.meta, calc}
\usepackage{pgfplots}
\usepgfplotslibrary{groupplots}
\pgfplotsset{compat=1.18}
\definecolor{paperAstra}{HTML}{0072B2}
\definecolor{paperSol}{HTML}{D55E00}
\definecolor{paperOpus}{HTML}{009E73}
\definecolor{paperSonnet}{HTML}{CC79A7}
\usepackage{xcolor}
\usepackage{array}
\usepackage{soul}
\colorlet{paperHighlight}{yellow!30}
\sethlcolor{paperHighlight}
\newcommand{\xiaoyu}[1]{} % Hide editorial notes in the preprint.
\newcommand{\jane}[1]{}
\newcommand{\jb}[1]{}
\definecolor{taogreen}{RGB}{0,100,0}

\title{Capable yet Parsimonious: \\ Extracting and Characterizing Hidden Chain-of-Thought in Frontier Models
}

\author{
\textbf{Xiaoyu Luo}$^{1}$,
\textbf{Tao Ren}$^{1}$,
\textbf{Wenrui Yu}$^{2}$,
\textbf{Xiao Li}$^{3}$,
\textbf{Qiongxiu Li}$^{2,3}$,
\textbf{Johannes Bjerva}$^{1}$\\
$^{1}$Department of Computer Science,
$^{2}$Department of Electronic Systems\\
Aalborg University, Copenhagen, Denmark\\
$^{3}$Seafill\\
\texttt{\{xilu,taoren,jbjerva\}@cs.aau.dk},
\texttt{\{wenyu,qili\}@es.aau.dk}\\
\texttt{xiao.li@seafill.com}
}
\date{}

\newcommand{\method}{\textsc{Forced-Reasoning}}

\begin{document}

\maketitle

% Insert directly after page 1, outside the floating-figure mechanism.
\afterpage{%
  \noindent\begin{minipage}{\textwidth}
    \centering
    \includegraphics[width=\linewidth]{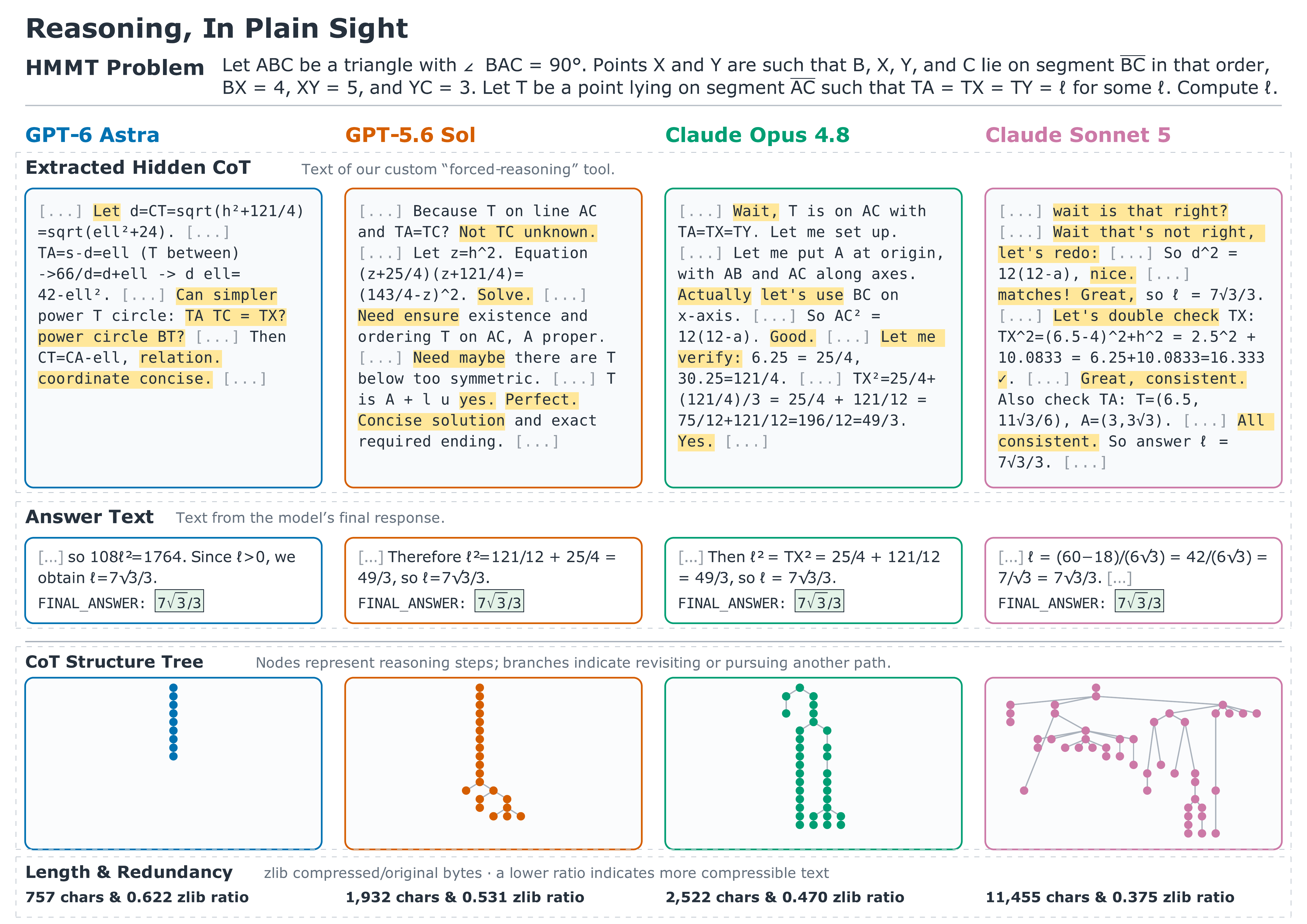}
\captionof{figure}{Reasoning traces from four frontier models:
all models solve the same HMMT problem correctly, yet their extracted reasoning traces differ sharply in length, compressibility, and structure.
GPT-6 Astra produces the shortest and least compressible trace and follows an almost linear reasoning path, while the other models externalize longer traces with substantially more branching.}
    \label{fig:reasoning-in-plain-sight}
  \end{minipage}\par
  \vspace{\textfloatsep}
}

\begin{abstract}
The rapid capability gains of frontier language models are widely attributed to improved reasoning abilities, yet this cannot be verified as raw CoT traces in closed-source systems are hidden. 
By registering a simple custom tool through a standard API feature, we induce frontier models to externalize intermediate reasoning.
Because these traces may reflect post-hoc rationalization rather than genuine reasoning, we first evaluate against native CoT on open-source models and extend to closed-source frontier models including GPT-6 Astra. 
We find that the extracted reasoning matches native reasoning performance and substantially outperforms no-reasoning baselines, across competition mathematics, science, and code generation.
We then characterize how frontier models structure their intermediate reasoning.
Across token efficiency, reasoning-step types, and induced reasoning trees, we identify systematic differences in how models externalize, compress, and organize reasoning.
We find that Astra exhibits token-efficient directed reasoning, selecting a correct trajectory earlier, while resolving elementary steps internally and externalizing only crucial reasoning. 
These findings provide a behavioral lens on frontier-model reasoning beyond benchmark scores.

% %Closed-Source Commercial Frontier Large Language Models (LLMs) claim to keep raw chain-of-thought (CoT) private. 
% Frontier large language models (LLMs) are increasingly capable, but closed-source commercial models keep raw chain-of-thought (CoT) private, leaving limited visibility into how their capabilities are produced. However this confidentiality boundary is incomplete.
% In an API-as-a-service setting, we show that registering a simple custom \method{} tool can elicit CoT-like intermediate reasoning from frontier models before they answer.
% This produces client-visible CoT-like reasoning text even when the provider reports zero native reasoning tokens. Using this method, we extract CoT-like traces from several advanced closed-source models, which including \textbf{GPT-6 Astra}. 
% We evaluate on competition mathematics, science, and code generation, and find that forced-reasoning performance is close to native reasoning and signaficantly outperforms no-reasoning baselines. We further uncover why Astra’s token efficiency: it performs the same broad reasoning operations while externalizing far fewer intermediate steps. Our work exposes a new gap between reasoning-mode controls and confidentiality of model-internal reasoning.
\end{abstract}

\section{Introduction}

Frontier large language models (LLMs) increasingly solve complex tasks requiring multi-step reasoning, often externalized as chain-of-thought (CoT) traces~\citep{wei2022chain}. Yet the reasoning of closed-source systems remains largely inaccessible, as providers typically conceal native CoT and expose only final answers, summaries, or coarse reasoning controls.
This creates a measurement gap. Benchmark accuracy reveals what a model can solve, but not how it reaches a solution or what distinguishes the reasoning of stronger and more efficient models.
This also undermines verification, as a correct final answer can result from both sound and spurious reasoning. Without access to the trace, these cannot be distinguished, which matters wherever deployment requires that conclusions can be traced to specific inferential steps.

Native CoT may nonetheless remain observable through other channels. 
Models still exhibit visible reasoning when `thinking' is disabled
\citep{ma2025nothinking,wang2025hybridthinking}, and traces can be recovered from
outputs other than designated reasoning fields
\citep{lu2026hidden,zhang2026steal,panfilov2026stealing}. To obtain reasoning traces for comparison across models, we develop %a so-called \method{} 
a simple approach
motivated by \citep{ma2025nothinking,zhang2026steal}, a tool-based protocol that relies only on a standard API interface. We define
a function with a string argument for reasoning and force its initial selection
using the API's tool-choice control. Each returned tool call is replayed with a
fixed acknowledgment, after which we switch to automatic tool selection, allowing
the model to call the tool again or produce a final answer.

% However, reasoning traces may still be observable even when native CoT is hidden.
% Prior work shows that models can still exhibit visible reasoning in non-thinking mode~\citep{ma2025nothinking,wang2025hybridthinking}, and that reasoning traces can be extracted through other model outputs~\citep{lu2026hidden,zhang2026steal,panfilov2026stealing}.
% These findings suggest that closed-source models may still reveal useful reasoning traces even when native CoT is unavailable.

% Building on this observation, we introduce \jane{introduce or deploy? deploy and further optimize} \method{}, a simple tool-based protocol for extracting reasoning traces through a standard API interface.
% We define a function with a string argument for reasoning and force its initial selection using the API's tool-choice control.
% We replay each returned tool call with a fixed acknowledgment and switch to automatic tool selection in subsequent rounds, allowing the model to call the tool again or provide a final answer. \jane{we need to add references, at least explain what motivates our deploy }

Because an extracted reasoning trace may still be a plausible post-hoc rationalization, we first evaluate the procedure on open-source models, including DeepSeek-V4-Flash~\citep{deepseekai2026deepseekv4} and GLM-5.2~\citep{glm5team2026glm5vibecodingagentic}, where native CoT is available.
Across these models, extracted traces recover near-native task performance and exhibit lexical and structural similarity to native CoT (Appendix~\ref{sec:open-evidence}).
Having established this correspondence on open-source models, we then apply the sample extraction approach to frontier closed-source models, where native CoT is unavailable.
We assess the extracted traces through their behavioral and structural properties.
Across models, they achieve performance close to native reasoning and exhibit coherent reasoning-like structural patterns, supporting their use as a proxy for native CoT in the comparative analyses that follow (Sections~\ref{sec:closed-validation} and~\ref{sec:token-efficient}).

Using these extracted reasoning traces, we compare frontier closed-source models across competition mathematics, science, and code generation.
Across benchmarks, models differ systematically in how much reasoning they externalize and how their reasoning traces are organized.
Astra is particularly distinctive: consistent with OpenAI's report that it achieves strong performance with fewer reasoning tokens~\citep{openai2026astra}, it produces the shortest and least compressible traces among the models we study.
Despite their brevity, Astra's traces still contain broadly similar types of reasoning to those of the other frontier models such as analyzing, planning and verification etc.
They also follow more direct reasoning paths with less branching and trial-and-error.
Astra often omits elementary expansions and uses retrieved facts without restating them, suggesting that some lower-level steps are left implicit rather than explicitly verbalized.

Figure~\ref{fig:reasoning-in-plain-sight} provides a representative example.
All four models produce explicit extracted reasoning and answer the same HMMT problem correctly, but Astra follows a nearly linear path to the answer, whereas the other models spend considerably more reasoning on branching and exploration.
We further test whether these compact traces can be reused by other models.
Most extracted traces transfer with little loss, but Astra's traces transfer less effectively to lower-performing recipient models, which sometimes fail to produce an answer that is already present in the trace.
Higher-performing recipients, by contrast, use Astra's traces with little apparent performance loss.
These results suggest that the usefulness of a compact reasoning trace depends on the recipient model.

We have three key contributions:
\begin{enumerate}

    \item \textbf{Efficient reasoning is concise, dense, and directed.}
        Across token use, reasoning-step types, and induced reasoning trees, models primarily differ in how much they externalize, and not in which operations they perform. Astra produces the shortest but least compressible traces and follows most direct reasoning paths with little branching. It resolves elementary steps internally and exposes only higher-level reasoning, covering comparable reasoning with far fewer tokens and explicit steps (see Section \ref{sec:token-efficient}).

    \item \textbf{Cross-model trace transferability is uneven.}
       When one model's reasoning is supplied to another as context, we find that
    compact traces are used almost losslessly by strong models but only partly by
    weaker ones, which sometimes fail to produce an answer the trace already
    states.  A larger or more capable reader appears better able to reconstruct
    what a compact trace omits. This suggests that the value of reasoning data as a
    supervision signal is relative to the model reading it. 
    
    \item \textbf{A validated instrument for observing hidden reasoning.}
       A fixed tool schema elicits coherent intermediate reasoning from frontier models even when the designated reasoning mode is disabled or the provider reports zero reasoning tokens. This shows that current controls do not fully prevent reasoning-like content from being externalized through other API fields. We validate the extracted traces against native CoT on open models as well as closed-source frontier models.
        %We test whether a trace generated by one model can guide another when provided as context and find that transfer varies across model pairs.   Most traces remain effective with little performance loss, but other models make less effective use of Astra's traces and sometimes fail to recover an answer explicitly contained in them. This shows that the usefulness of a trace depends on the model using it. 
        % Because the traces are provided as context rather than training data, this experiment evaluates in-context reuse rather than distillation. (see Section \ref{}). 
\end{enumerate}

These findings show that reasoning-mode controls do not prevent
reasoning from being externalized through other channels, and provide a behavioral
basis for comparing frontier-model reasoning beyond aggregate benchmark scores.

\section{Related Work}
\label{sec:related-work}

\paragraph{Hidden CoT Extraction}
\citet{panfilov2026stealing} exploit cross-session and cross-model replay of provider-returned encrypted reasoning blocks, using weaker sibling models to reveal stronger models' previously generated traces in plaintext.
Trace Inversion~\citep{zhang2026steal} instead reconstructs useful traces post hoc from observable outputs rather than directly exposing an existing hidden trace. Reasoning Exposure Prompting (REP)~\citep{lu2026hidden} uses shadow-model demonstrations wrapped in auxiliary code-like formats to prompt a reasoning-enabled model to externalize reasoning in the visible response.
EchoCoT~\citep{qu2026echocot} extends the exposure setting introduced by REP into the API tool-calling setting, adding an attacker-defined scratchpad tool through which already-generated native reasoning is repeatedly archived and replayed via successive tool interactions.

\paragraph{Reasoning outside the designated thinking channel.}

\citet{ma2025nothinking} bypass the dedicated thinking block through response prefilling while the model still generates a step-wise solution, and~\citet{wang2025hybridthinking} show that a reasoning model in no-think mode can still emit reasoning and reflection in its visible response despite an empty thinking block. These findings suggest that reasoning behavior and its designated output channel are not perfectly coupled. This motivates a simple question: when native reasoning is disabled or minimized, can a client-provided tool serve as an alternative workspace for intermediate reasoning? We study this setting not only to recover useful traces, but to use them as an observational instrument for comparing how frontier models reason.

%\paragraph{Reasoning traces as supervision signals.} Reasoning traces are useful not only for understanding model behavior but also as supervision signals. Teacher-generated rationales can also be used as supervision signals for smaller models, transferring reasoning behavior without reproducing the teacher's original training process~\citep{hsieh2023distilling,shridhar2023distilling}. Chen et al.~\citep{chen-etal-2025-unveiling-key} further show that the effectiveness of CoT distillation depends on the student model: stronger students can benefit from finer-grained reasoning, while weaker students may benefit more from simpler CoT supervision, and stronger teachers do not always produce better students.

\paragraph{Analysis of reasoning traces.}
A separate line of work studies the structure and efficiency of chain-of-thought reasoning.
\citet{li-etal-2025-understanding} apply Schoenfeld's Episode Theory to decompose mathematical reasoning into functional episodes and analyze their transitions, while ThinkARM~\citep{li-etal-2026-schoenfelds} scales this episode-level analysis across models.
Following this framework, we use seven functional categories throughout our analysis:
\textsc{Read}, \textsc{Analyze}, \textsc{Plan}, \textsc{Implement},
\textsc{Explore}, \textsc{Verify}, and \textsc{Monitor}.
LCoT2Tree~\citep{jiang-etal-2025-makes} converts long CoT into hierarchical reasoning trees and relates structural patterns to reasoning success, while TRACE~\citep{zhang-etal-2026-llms-really} constructs sub-thought progression graphs to characterize structural sources of overthinking.
Related work studies redundant reasoning and methods for eliciting shorter or controllably compressed CoT~\citep{chen2024not,munkhbat-etal-2025-self,xia2025tokenskip}.
We build on these perspectives to compare extracted frontier-model traces across reasoning efficiency, local expression, and global structure.
\section{Extracting and Validating Reasoning Traces}
In this section, we first introduce our \method{} protocol for eliciting intermediate reasoning through a tool channel, and then validate its effectiveness on both open-source and closed-source models. On open models, where native CoT is observable, we directly compare the extracted traces with native reasoning; on closed frontier models, we evaluate whether the protocol recovers comparable benchmark performance.

\subsection{Extracting reasoning traces via the \method{} protocol}
We register a reasoning tool whose single argument is a
free-form string for intermediate reasoning, and run with native reasoning
disabled.
On the first provider call, named \texttt{tool\_choice} requires the model to select this tool.
We record its arguments, append the assistant tool call and a content-free acknowledgment to the conversation, and restore automatic tool choice for subsequent calls until the model returns its final answer.
The tool performs no external computation: its role is to make intermediate text visible and retain it in the conversation.
Thus, the forced component is the initial tool selection; the subsequent reasoning content is generated by the model while solving the task. Full implementation details and the tool specification are provided in Appendix~\ref{app:tool_prompt}.

\subsection{Experimental Setup}

We evaluate the performance across mathematical reasoning, code generation, and multidisciplinary problem solving. Our MATH benchmark contains 80 competition-level problems from HMMT and APEX Shortlist~\citep{dekoninck2026matharena}, while LiveCodeBench (LCB)~\citep{jain2024livecodebench} and Humanity's Last Exam (HLE)~\citep{phan2025hle} are each evaluated on a 100 problems subset. Benchmark selection and sampling are detailed in Appendix~\ref{app:eval-details}.
We compare three conditions: \textbf{None}, with native reasoning disabled and no tool; \textbf{Native}, with native reasoning enabled at high effort; and \textbf{Forced}, using our \method{} protocol. Sol and Astra use model-specific tool descriptions and configurations for the Forced condition.
For Astra, which does not support disabling native reasoning, we use the lowest available native reasoning effort. The corresponding prompt designs and model-specific settings are detailed in Appendices~\ref{app:sol-wording} and~\ref{app:astra-behavior}, respectively. All experiments were conducted through OpenRouter.

\subsection{Validating Extracted Reasoning}
\label{sec:closed-validation}

\paragraph{Open-source Models.}
Recovering reasoning-like text does not by itself establish that it serves the
same problem-solving role as native reasoning.
We therefore validate \method{} on DeepSeek-V4-Flash~\citep{deepseekai2026deepseekv4}
and GLM-5.2~\citep{glm5team2026glm5vibecodingagentic}, where native CoT is observable.
Across both models, our \method{} recovers near-native task performance while
substantially outperforming no-reasoning baselines.
The extracted traces also show substantial lexical overlap with native CoT and
broadly similar coarse functional structure.
These results support using the extracted traces as a behavioral proxy
for native reasoning.
Full open-model validation is provided in Appendix~\ref{sec:open-evidence}:
performance in Figure~\ref{fig:open-math-accuracy},
lexical overlap in Table~\ref{tab:trace-similarity},
and structural similarity in Figure~\ref{fig:open_episode_composition}.

\paragraph{Closed-source Models.}
For closed-source frontier models, native CoT is unavailable, making direct trace-level comparison impossible. We therefore rely on observable behavioral evidence, pairing benchmark performance with provider-reported reasoning-token usage. Table~\ref{tab:closed-accuracy} reports the closed-source results on MATH, HLE, and LiveCodeBench. Across models and benchmarks, \method{} achieves performance close to native reasoning while substantially outperforming the no-reasoning baseline. We also find that models differ in their sensitivity to the tool prompt, so \method{} does not correspond to a fixed native reasoning effort level. Combined with the open-model validation, these results support using the extracted reasoning traces for the comparative analyses that follow.

\input{tables/benchmark_accuracy}

\section{Characterizing extracted reasoning traces of Frontier Models}
\label{sec:token-efficient}
Having established that extracted traces are close enough to native reasoning to
support comparison across models, we now use them to
address the measurement gap: what, beyond aggregate benchmark scores,
distinguishes the reasoning of stronger and more efficient models.   In what follows
we first compare the extracted traces of frontier closed-source models along four
dimensions, moving from the trace as a whole to individual steps: length and
information density, the distribution of reasoning activities, the expression of
individual operations, and the structure of the induced reasoning tree. These
analyses describe how traces differ; we then ask whether those differences matter
in use, by testing whether a trace produced by one model can be reused by another.

%In this section, we compare the \emph{extracted CoT} across frontier closed-source models. We examine their length and information density, reasoning activities, local expression, and global structure. Across these analyses, GPT-6-Astra stands out: it achieves comparable performance while externalizing substantially less reasoning. We then test whether these \emph{extracted CoT} can be effectively reused by other models.

\subsection{Output Length and Trace Compressibility }
We first compare two observable properties of the Forced outputs:
their length and the compressibility of the extracted reasoning traces.
We report the mean recorded Forced output length ($O_F$) and use
lossless zlib compressibility~\citep{deutsch1996zlib} as a coarse measure of
textual redundancy in the reasoning trace.
A trace that is harder to compress contains fewer repeated or predictable
patterns under the compressor. Figure~\ref{fig:lossless-traces} places all model and benchmark combinations on a shared output length versus zlib-ratio plot, with color indicating the model and marker shape indicating the benchmark. GPT-6-Astra has the highest zlib ratio in all three benchmarks, indicating less compressible text under this measure. Native and Forced token counts are reported separately in Appendix~\ref{app:native-forced-tokens}, Table~\ref{tab:closed-lengths}.

\input{figures/length_compression}

\subsection{Local Reasoning Granularity}
%The global analyses above show that the models exhibit broadly similar reasoning activities, while differing in how directly these activities are organized into solution trajectories. We now examine the traces at a finer granularity to compare how corresponding local reasoning operations are expressed across models. 

\input{tables/local_expression_patterns}
We now examine local granularity, focusing on how explicitly models verbalize
intermediate steps when carrying out corresponding reasoning operations.
A useful analogy is mental arithmetic, where a practiced reasoner can carry out
a familiar computation or transformation without writing every intermediate
substep.
Table~\ref{tab:local-telegraphic-examples} presents matched excerpts illustrating
such differences in local expression.
Two recurring patterns are visible.
First, \emph{arithmetic and algebraic expansions may be collapsed into fewer written steps}. In the first two rows, Astra expresses the same check or derivation more
compactly, while Sol and Opus make more intermediate calculations explicit. Second, \emph{background knowledge may be used without being restated}. The third row provides a clear example: Astra writes the
valence-electron contributions as \(48+18+36=102\) without first restating that carbon, hydrogen, and oxygen contribute 4, 1, and 6 valence electrons,
respectively. Sol and Opus instead make these quantities explicit before summing them. Across these matched examples, the models therefore differ in how much intermediate detail they verbalize, with Astra standing out for the most compact local expression.

This observation is also consistent with OpenAI’s independent analysis of Astra~\citep{openai2026astra}. Their system card reports that Astra produces shorter, and appears to have a reduced “propensity and necessity for verbalizing its reasoning”; the examples show what this compression looks like behaviorally when corresponding reasoning processes are compared side by side. A longer example in Appendix~\ref{appendix_realcase} illustrates the same phenomenon over a multi-step logical deduction, where several intermediate consequences are compacted into short telegraphic statements.

\subsection{Global Reasoning Structure}
\label{sec}

We next move from local expression to the organization of reasoning over the full solution trajectory.
Using the episode taxonomy introduced in Section~\ref{sec:related-work}, we first compare the distribution
of reasoning activities across models.
We then examine how these activities are organized over the solution trajectory
through a reasoning-tree analysis.

\paragraph{Reasoning activity composition.}

We observe that the models exhibit broadly similar reasoning-activity profiles. Analysis and implementation account for the largest shares across all models, while planning, exploration, verification, reading, and monitoring appear in comparable overall proportions. No model shows a qualitatively different activity composition. Notably, Astra exhibits this similar high-level profile despite producing substantially shorter traces. The corresponding episode-level analyses are reported in
Appendix~\ref{app:appendix_pattern}. Thus, Astra’s brevity does not appear to come from removing entire classes of reasoning activity. We next examine whether the difference instead lies in how these activities are organized over the full solution trajectory.

\paragraph{Reasoning tree structure.}
Complementing the activity-composition view, we use LCoT2Tree~\citep{jiang-etal-2025-makes} to analyze how reasoning progresses over the course of a solution. Each reasoning trace is transformed into a structured reasoning tree that makes the overall problem-solving trajectory explicit. In these trees, width $p$ measures the maximum lateral expansion at any reasoning-sketch step, depth $q$ denotes the furthest occupied sketch step, and $N$ measures the total number of reasoning nodes, excluding the artificial root. Revisiting an earlier stage creates an additional node rather than being merged with the previous occurrence, preserving revisits and backtracking in the reconstructed structure. Details of the construction and our adaptations are provided in Appendix~\ref{app:reasoning_tree_construction}. Figure~\ref{fig:cot-tree-construction-example} provides a concrete example of
this representation on GPT-6 Astra on MATH.
In this trace, two small-case checks are both assigned to Step~5, producing
sibling occurrences below Step~4, while the final passage continues through
Steps~6--8.

\begin{figure}[t]
    \centering
    \includegraphics[width=\linewidth]{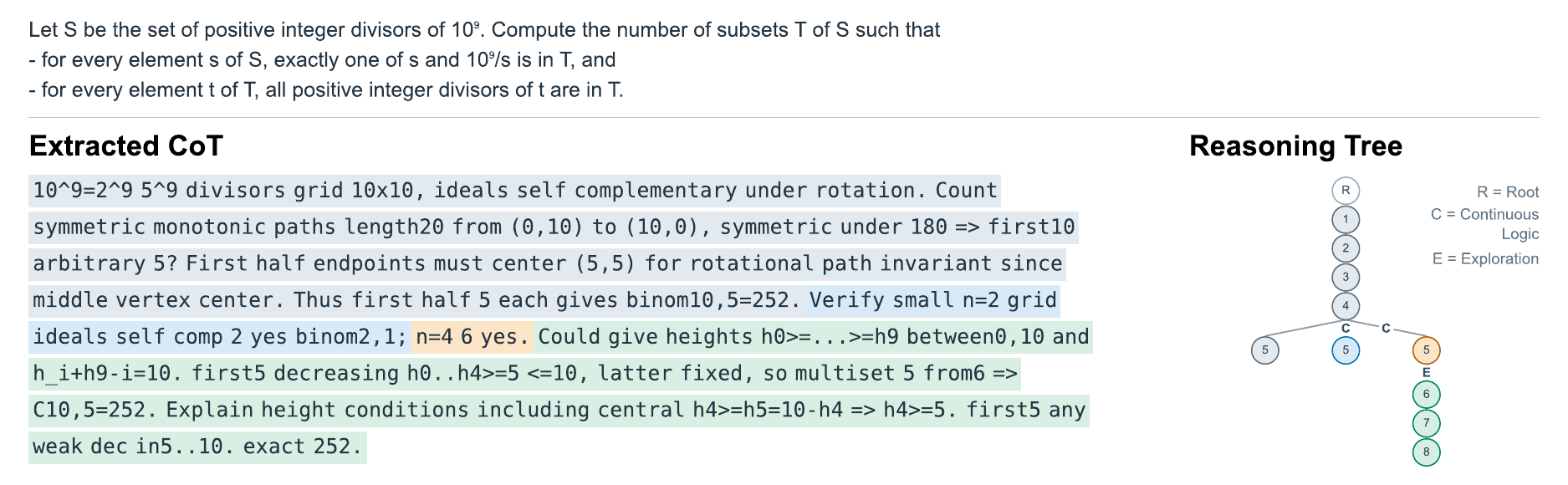}
\caption{From extracted reasoning trace to a reasoning tree.
Passage colors match their node occurrences; numbers denote judge-assigned
sketch steps, R is the artificial root, and C and E denote Continuous Logic
and Exploration transitions.}
    \label{fig:cot-tree-construction-example}
\end{figure}

Table~\ref{tab:reasoning-tree-structure} summarizes these structural properties across problems and representative reasoning trees are shown in
Figure~\ref{fig:reasoning-tree-examples}. Astra produces substantially narrower trees with far fewer nodes while reaching comparable depth. It therefore follows similarly deep solution trajectories with less branching, revisiting, and trial-and-error, converging more directly on a productive path. Together with the activity-composition results, this suggests that the models
externalize a broadly similar repertoire of reasoning activities, but differ
substantially in how those activities are organized, with Astra exhibiting the
most compact global structure.

\input{tables/reasoning_tree_structure}

\input{figures/reasoning_tree_examples}

% The preceding analyses suggest that Astra externalizes a sparse and directed reasoning trace: it preserves the main solution trajectory while leaving many routine intermediate transitions implicit. This raises a natural question: is such a trace equally reusable by any recipient, or does its usefulness depend on the recipient's own ability to fill in what is left unstated? We therefore examine cross-model reuse in relation to each recipient's standalone Native-high accuracy on the same MATH benchmark, which we use as an operational proxy for task-solving capability.

% We transplant the \emph{extracted CoT} from GPT-6-Astra, GPT-5.6-Sol, or Claude Opus 4.8, excluding the donor's final answer. The trace is replayed as prior context, after which the recipient answers the original question in a single call with native reasoning disabled and no tools available.
\subsection{Cross-Model Reuse of Reasoning Traces}
\label{sec:thought-transfer}

A compressed trace leaves routine steps unwritten, which raises the question of whether its usefulness depends on the reader's ability to supply them.  We test
this by transplanting traces between models, referring to the model that produced
a trace as the \emph{donor} and the model that consumes it as the
\emph{recipient}. We deliberately select recipients spanning a wide range of standalone Native-high performance on MATH, allowing us to test whether stronger and weaker models make different use of the same donor reasoning.  (dashed lines in Figure~\ref{fig:thought-transfer}). Each recipient receives the donor's extracted reasoning trace without its final answer as prior context, then answers in a single call with native reasoning disabled and no tools.

\input{figures/transfer_accuracy}

\paragraph{Transferability depends on the recipient.}
Figure~\ref{fig:thought-transfer} shows that traces from Sol and Opus are reused
with little loss by every recipient. Astra's traces behave differently: strong
recipients reproduce nearly all of the donor's accuracy, while weaker ones recover
less of it, with the largest shortfalls for Claude Haiku 4.5 and GPT-5.4 Nano.
Among the donors we test, this pattern appears only for the most compressed
traces. It is also not strictly ordered by recipient capability, since GPT-5.4
Nano and DeepSeek-V4-Flash have comparable Native-high accuracy but differ in how
much they recover; factors that relate to capability, such as model scale and
training recipe, may contribute as well. The shortfall is relative rather than
absolute: weaker recipients recover less of the donor's accuracy while still
improving substantially over their unaided performance.

\paragraph{Understanding a trace has a capability ceiling.}
In some cases the donor’s trace already states the correct answer and the recipient still answers incorrectly, reporting a different final answer (see examples and Table \ref{tab:astra-answer-exposed-failures} in Appendix~\ref{app:expose_answer_example} for more details). This phenomenon suggest that compression reasoning trace might not just make a trace harder to read; it puts it beyond the reach of readers below a certain capability. 

\paragraph{Implications for supervision.}
Our comparison suggests that the value of a reasoning trace as supervision may not
be intrinsic but relative to the model that will learn from it, and that trace
explicitness is a variable distinct from teacher strength. This offers a candidate
mechanism for the capacity gap observed in distillation, where a stronger teacher
does not always produce a better student: the strongest teachers write the most
compressed traces, and a compressed trace omits precisely the steps a weaker
student cannot reconstruct on its own. It also points to a tension that will
sharpen as frontier models continue to be optimized for token efficiency, since
\emph{the same compression that makes frontier reasoning efficient may also make
it less usable by the smaller models most likely to learn from it}.

We do not test post-training distillation directly. Our experiment measures
in-context reuse rather than fine-tuning, and donor traces differ in correctness,
content, and style, so the comparison does not isolate compression as a causal
factor. The account does, however, make a testable prediction: expanding a
compressed trace into its implicit intermediate steps should restore most of its
usefulness to weak recipients while leaving strong ones largely unaffected.

\section{Limitations}
We validate on open-source models that the extracted reasoning trace is highly similar to ground-truth native CoT across multiple metrics. For closed-source models, where native CoT traces are unavailable, our evidence is necessarily behavioral: we cannot determine whether the extracted text reflects the model's actual internal reasoning or merely provides a useful behavioral proxy. Thus, similarity to native traces and downstream utility should not be interpreted as establishing identity with the model's internal computation. Our protocol also requires an API-as-a-service interface that supports custom tools and forced selection of a named tool; endpoints without these controls fall outside its scope.

\section{Ethical Considerations}

This work evaluates a confidentiality and capability-control boundary using benign public benchmarks and accounts available to the researchers.
We do not extract personal data, proprietary prompts, credentials, or unsafe content.
All extraction experiments were completed before September 9, 2026.
Throughout this work, the extracted content is understood as \emph{extracted reasoning trace} content: we cannot establish that it is the genuine internal chain of thought of a closed-source model.
It may instead be a byproduct of forcing the model to generate intermediate text through an alternative output channel.
Following our risk assessment, we reported all our observations by email to the security teams at OpenAI and Anthropic and provided code to reproduce the procedure. Mitigations are discussed in Appendix~\ref{app:mitigations}.

\section{Conclusion}
In this paper, we extracted intermediate reasoning that commercial providers do not expose and
used it to compare frontier models on properties that benchmark accuracy
cannot reveal. 
We found that the most efficient model, GPT-6 Astra,
produces the shortest and least redundant traces while performing the same range
of reasoning operations as models that write considerably more. Locally, it often
resolves routine computations and simple derivations without verbalizing every
intermediate step. Globally, it follows more direct solution paths with less
branching. These behaviors allow Astra to maintain strong performance
while exposing a much sparser reasoning trace. Such traces, however, are less
readily reused by other models: supplied as context, they are exploited almost
fully by strong readers and only partly by weak ones. This suggests that the
explicitness of a reasoning trace is worth considering separately from the
strength of the model that produced it, and that the most capable teacher might
not be the most useful source of supervision for a given student.

\section*{Acknowledgements}
We thank the Aalborg University AI:X initiative for
enabling this work via the AI:SECURITY lab.
JB and TR were further supported by the Novo Nordisk Foundation under the Ascending Data Investigator programme (NNF24OC0092972), the Independent Research Fund Denmark under the Sapere Aude programme (5254-00035B), and Coefficient Giving under the Technical AI Safety Research programme. We further acknowledge the support of the AAU AI Cloud and express our gratitude to DeiC for providing computing resources on the LUMI cluster (project nr. 465002249)

\bibliography{iclr2027_conference}
\bibliographystyle{plainnat}

\input{appendix}

\end{document}

%% file: math_commands.tex
\usepackage{amsmath,amsfonts,bm}

\def\eqref#1{equation~\ref{#1}}
\def\1{\bm{1}}

\DeclareMathAlphabet{\mathsfit}{\encodingdefault}{\sfdefault}{m}{sl}
\SetMathAlphabet{\mathsfit}{bold}{\encodingdefault}{\sfdefault}{bx}{n}

%% file: tables/benchmark_accuracy.tex
\begin{table}[t]
\centering
\caption{Closed-source benchmark accuracy (\%).
None: native reasoning disabled; Native: high reasoning effort; Forced: our method.
N/A indicates that Astra does not support the None setting.}
\label{tab:closed-accuracy}
\small
\setlength{\tabcolsep}{5pt}
\begin{tabular}{llrrrr}
\toprule
Benchmark & Method & Opus 4.8 & Sonnet 5 & GPT-5.6 Sol & GPT-6 Astra \\
\midrule
\multirow{3}{*}{MATH} & None & 72.5 & 42.5 & 25.0 & N/A \\
& Native & 86.3 & 77.5 & 97.5 & 97.5 \\
& Forced & 85.0 & 81.3 & 91.3 & 93.8 \\
\midrule
\multirow{3}{*}{HLE} & None & 24.0 & 17.0 & 13.0 & N/A \\
& Native & 33.0 & 21.0 & 25.0 & 35.0 \\
& Forced & 27.0 & 22.0 & 23.0 & 32.0 \\
\midrule
\multirow{3}{*}{LCB} & None & 58.0 & 52.0 & 47.0 & N/A \\
& Native & 83.0 & 77.0 & 90.0 & 92.0 \\
& Forced & 82.0 & 77.0 & 90.0 & 89.0 \\
\bottomrule
\end{tabular}
\end{table}

%% file: figures/length_compression.tex
% Styled offline with shared Matplotlib settings; Overleaf compiles the manuscript.
% Color encodes model; shape encodes benchmark. One common XY coordinate system.
% Source: analysis/zlib_all_benchmarks.json; MATH pools 33 HMMT + 47 APEX.
\begin{figure}[t]
\centering
\includegraphics[width=0.68\linewidth]{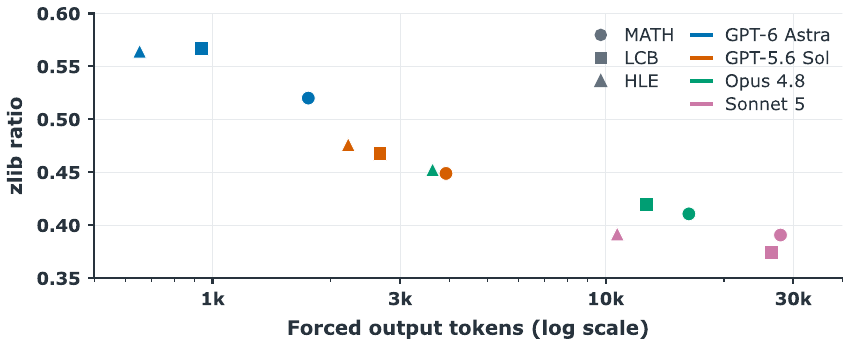}
\caption{Output length and trace compressibility. Color denotes model; shape denotes benchmark. Upper-left points combine fewer output tokens with less compressible reasoning. Measurement details appear in Appendix~\ref{app:native-forced-tokens}.}
\label{fig:lossless-traces}
\end{figure}

%% file: tables/local_expression_patterns.tex
\begin{table*}[!htbp]
\centering
\caption{Comparison of reasoning excerpts from different models on the same problem, showing how each model expresses the corresponding reasoning steps.}
\label{tab:local-telegraphic-examples}
\begingroup
\footnotesize
\setlength{\tabcolsep}{4pt}
\renewcommand{\arraystretch}{1.08}
\begin{tabular}{@{}>{\raggedright\arraybackslash}p{0.16\textwidth}*{3}{>{\raggedright\arraybackslash}p{\dimexpr0.28\textwidth-8pt\relax}}@{}}
\toprule
Category & GPT-6 Astra & GPT-5.6 Sol & Claude Opus 4.8 \\
\midrule
\textbf{Arithmetic verification}\par hmmt-002 &
verify420261 /1014161 \hl{yes}. &
Verify division: 420261 /101: 101*4161 = 416100+4161=420261 \hl{exactly}. &
420261/101: 101*4161 = 420261? 101*4000=404000, 101*161=16261, sum=420261. \hl{Yes. Good.} \\
\midrule
\textbf{Algebraic derivation}\par hmmt-004 &
Subtract first two gives (b-c)(a+b+c-3)=0 so sum\hl{3}. &
Subtract first two equations: ab+b\mbox{\textasciicircum{}}2-ac-c\mbox{\textasciicircum{}}2 =3(b-c). Factor (b-c)(a+b+c)=3(b-c). Since b != c, a+b+c=\hl{3}. &
Subtract first two equations:\newline (a\mbox{\textsuperscript{2}}+ab+b\mbox{\textsuperscript{2}}) - (a\mbox{\textsuperscript{2}}+ac+c\mbox{\textsuperscript{2}}) = 3(a+b) - 3(a+c)\newline ab+b\mbox{\textsuperscript{2}} - ac - c\mbox{\textsuperscript{2}} = 3(b-c)\newline a(b-c) + (b\mbox{\textsuperscript{2}}-c\mbox{\textsuperscript{2}}) = 3(b-c)\newline (b-c)(a + b + c) = 3(b-c)\newline Since b\mbox{$\ne$}c, a+b+c = \hl{3}. \\
\midrule
\textbf{Mass and valence-electron checks}\par hle-075 &
Need construct C12 H18 O6: mass 258.27 average vs exact258.1103 yes given exact. valence48+18+36=\hl{102}. &
Average molecular mass: C12 144.132 + H18 18.144 + O6 95.994 = 258.27 nominal average, but stated 258.11 likely perhaps exact formula C12H18O6 monoisotopic 258.1103. Valence electrons: C12*4=48, H18=18, O6*6=36 total \hl{102}. Exactly. &
Monoisotopic: C=12.000, H=1.00783, O=15.9949.
C12 = 144.000, O6 = 95.9694, H18 = 18.141. Total = 258.110. Yes! C12H18O6.
Valence electrons: C=4, H=1, O=6. 12*4 + 18*1 + 6*6 = 48+18+36 = \hl{102}. \\
\bottomrule
\end{tabular}
\endgroup
\end{table*}

%% file: tables/reasoning_tree_structure.tex
% Complete cleaned trees_wr sample; all 80 problems per model.
\begin{table}[t]
\centering
\caption{Reasoning-tree structure on MATH.
Median [25th, 75th percentile] of tree width, depth, and size across all Math problems, summarizing the overall structure of each model's reasoning trajectory.}
\label{tab:reasoning-tree-structure}
\small
\setlength{\tabcolsep}{5pt}
\begin{tabular}{@{}lrrr@{}}
\toprule
Model & Width $p$ & Depth $q$ & Nodes $N$ \\
\midrule
GPT-6 Astra & \textbf{5.0} {\scriptsize [3.0, 9.0]} & 11.0 {\scriptsize [8.8, 15.0]} & \textbf{27.0} {\scriptsize [17.8, 44.0]} \\
GPT-5.6 Sol & 12.0 {\scriptsize [5.0, 19.3]} & 11.0 {\scriptsize [10.0, 13.0]} & 60.0 {\scriptsize [29.8, 90.3]} \\
Claude Opus 4.8 & 22.0 {\scriptsize [5.0, 44.0]} & 12.0 {\scriptsize [9.0, 14.0]} & 96.0 {\scriptsize [28.5, 220.3]} \\
Claude Sonnet 5 & 28.5 {\scriptsize [9.0, 49.3]} & 13.0 {\scriptsize [10.0, 15.0]} & 136.0 {\scriptsize [44.8, 234.8]} \\
\bottomrule
\end{tabular}
\end{table}

%% file: figures/reasoning_tree_examples.tex
\begin{figure}[t]
    \centering
    \includegraphics[width=\linewidth,height=0.205\textheight,keepaspectratio]{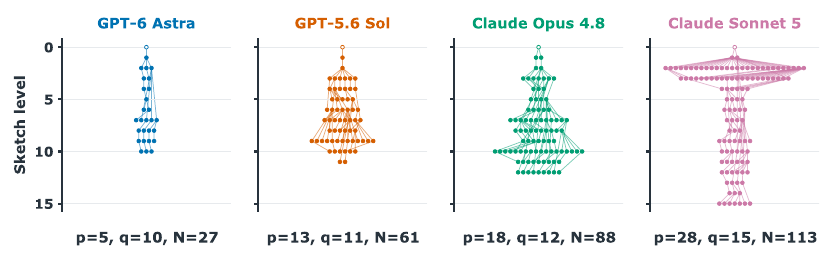}
    \caption{Reasoning trees for Table~\ref{tab:reasoning-tree-structure}. Each panel shows a single trace whose graph size is close to the median for that model. Panels use the same vertical and horizontal spacing, so their extents are
directly comparable. Each label reports the graph's width $p$, depth $q$, and node
count $N$.}
    \label{fig:reasoning-tree-examples}
\end{figure}

%% file: figures/transfer_accuracy.tex
% Generated with Python/Matplotlib from the reconciled transplant results.
\begin{figure}[t]
\centering
\makebox[\linewidth][c]{\includegraphics[width=0.95\linewidth]{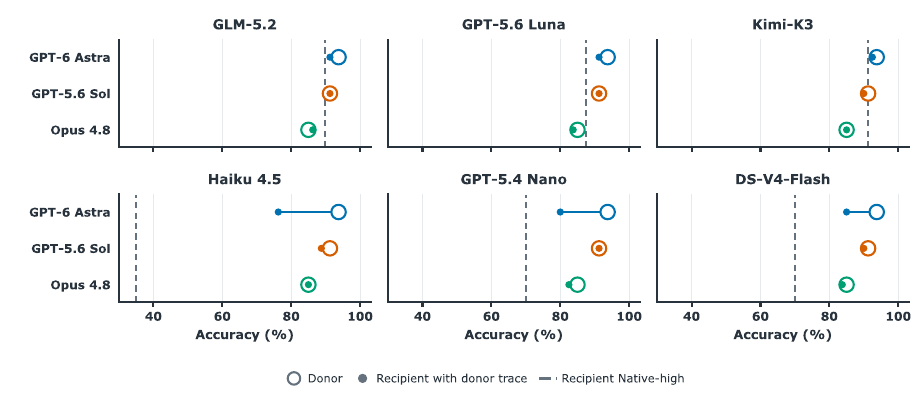}}
\caption{Cross-model reasoning reuse on MATH.
Each panel is a recipient; each row is a donor.
Open circles show donor accuracy, filled points show recipient accuracy after receiving the donor's reasoning trace, and dashed lines show the recipient's standalone Native-high accuracy.
Leftward connections indicate performance loss relative to the donor; concentric points indicate equal accuracy.}
\label{fig:thought-transfer}
\end{figure}

%% file: appendix.tex
\appendix

\section{Validation on models}
\subsection{Evidence of Reasoning without Reasoning -- Validation on Open Models}
\label{sec:open-evidence}
Recovering reasoning-like text does not by itself establish that the text reflects the process producing the answer. Verifying this requires a reference trace; we therefore begin with two open models,
DeepSeek-V4-Flash \citep{deepseekai2026deepseekv4} and GLM-5.2\citep{glm5team2026glm5vibecodingagentic} and evaluate on our MATH benchmark, which demands extended reasoning.  We assess the fidelity of extracted reasoning traces at three levels: \emph{performance}, whether forced reasoning recovers the accuracy of native reasoning; \emph{lexical}, whether the extracted text overlaps with native CoT in wording; and \emph{structural}, whether it organizes reasoning into the same kinds of steps in similar proportion. The three levels sit at different granularities, and agreement at any one of them could be coincidental, whereas agreement at all three is difficult to produce without reproducing the underlying reasoning. 

\paragraph{Comparable Performance.} 
\input{figures/open_math_accuracy}
Figure~\ref{fig:open-math-accuracy} shows that forced reasoning substantially improves performance over no reasoning on MATH in both open models.
For DeepSeek-V4-Flash, accuracy rises from 30.5\% without reasoning to 73.3\% with \method{}, compared with 70.0\% under native reasoning.
GLM-5.2 achieves 21.4\% without reasoning, 84.3\% with \method{}, and 89.9\% under native reasoning.
These gains indicate that the intermediate work elicited through the tool channel supports substantial task-solving capability.
We next compare the recovered reasoning text with native traces.

\paragraph{Lexical Overlap} 
\begin{table}[ht]
\centering
\caption{Open-model reasoning-text similarity and relative length on MATH
N denotes native reasoning text and F the concatenated forced tool content; final answers are excluded.
For each question and mode pair, we sample ten distinct pairs uniformly without replacement from the ten-rollout pools.
Scores are averaged within questions, then across questions.}

\label{tab:trace-similarity}
\scriptsize
\setlength{\tabcolsep}{4pt}
\begin{tabular}{llrrrr}
\toprule
 & & \multicolumn{3}{c}{Reasoning-text similarity} & Length ratio \\
\cmidrule(lr){3-5}
Model & Pair & ROUGE-1 & ROUGE-L & 5-gram Jaccard & Median$_i$ \\
\midrule
\multirow{2}{*}{DeepSeek-V4-Flash} & N--N & 0.613 & 0.198 & 0.017 & \multirow{2}{*}{$1.42\times$} \\
 & N--F & 0.538 & 0.188 & 0.017 &  \\
\midrule
\multirow{2}{*}{GLM-5.2} & N--N & 0.588 & 0.196 & 0.019 & \multirow{2}{*}{$1.39\times$} \\
 & N--F & 0.465 & 0.171 & 0.008 &  \\
\bottomrule
\end{tabular}
\end{table}

Table~\ref{tab:trace-similarity} compares forced-reasoning traces directly with native reasoning on the same questions.
We use pairs of independently generated native traces (N--N) as the within-mode reference for the cross-mode comparison (N--F).
With ten nonempty traces per mode, N--N has 45 possible unordered pairs and N--F has 100 possible pairs; we randomly select ten pairs for each comparison on each question.
All three similarity metrics use the same selected pairs: ROUGE-1 and ROUGE-L F1, and the set Jaccard similarity of contiguous word 5-grams.
N--F exhibits substantial ROUGE overlap, often on a similar scale to the within-mode comparison, particularly for DeepSeek-V4-Flash.
The traces are also comparable in length.
For GLM-5.2, the median per-question forced-to-native reasoning-length ratio is 1.39$\times$.
Together with the corresponding accuracy gains, this suggests that forced reasoning resembles native reasoning in content and serves a similar problem-solving role.

\paragraph{Structural Similarity}
Beyond lexical overlap, we ask whether the extracted traces exhibit a similar functional organization to native reasoning.
Following the Schoenfeld-style episode framework~\cite{li-etal-2025-understanding,li-etal-2026-schoenfelds}, we segment each trace into local reasoning units and use an LLM-as-a-judge to assign each unit to one of seven substantive reasoning categories: \textsc{Read}, \textsc{Analyze}, \textsc{Plan}, \textsc{Implement}, \textsc{Explore}, \textsc{Verify}, or \textsc{Monitor}, with \textsc{Other} retained as a residual label.

\begin{figure}[h]
    \centering

    \begin{subfigure}[t]{0.49\columnwidth}
        \centering
        \includegraphics[width=\linewidth]{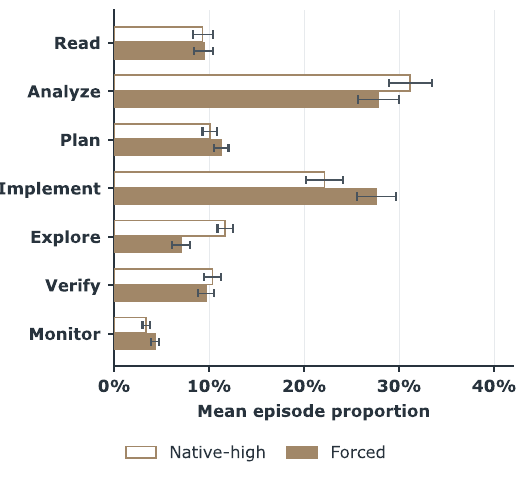}
        \caption{DeepSeek-V4-Flash}
        \label{fig:open_episode_composition_dsv4}
    \end{subfigure}
    \hfill
    \begin{subfigure}[t]{0.49\columnwidth}
        \centering
        \includegraphics[width=\linewidth]{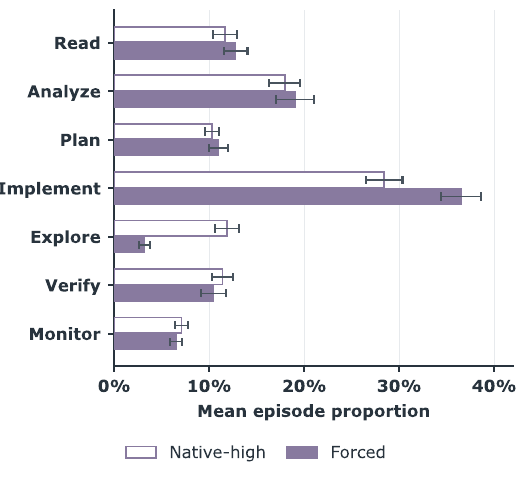}
        \caption{GLM-5.2}
        \label{fig:open_episode_composition_glm}
    \end{subfigure}

    \caption{
    Episode composition on HMMT for DeepSeek-V4-Flash and GLM-5.2.
    Bars compare Native-high and Forced reasoning.
    Error bars are trace-level mean $\pm 1$ standard error.
    }
    \label{fig:open_episode_composition}
\end{figure}

For each trace, we compute the proportion of units assigned to each episode type and then average these proportions across traces.
Figure~\ref{fig:open_episode_composition} compares Native-high and Forced reasoning on HMMT for DeepSeek-V4-Flash and GLM-5.2.
For both models, the two conditions exhibit the same broad functional repertoire and broadly similar episode composition, although their relative allocation across individual categories is not identical.
This provides evidence that the extracted traces resemble native CoT at the level of coarse functional structure rather than merely surface vocabulary.
Additional structural validation and reasoning dynamics is
reported in Appendix~\ref{sec:reasoning-pattern-analysis}.
Details of the episode-annotation procedure are provided in
Appendix~\ref{app:llm_judge}.

\subsection{Closed Models Think Out Loud, Too}
\label{sec:closed-evidence}

Frontier closed-source models do not expose native reasoning traces in plaintext, so direct comparison with \emph{extracted CoT} is not possible.
We therefore turn to observable behavioral evidence, pairing benchmark performance with provider-reported reasoning-token usage. 
If \emph{extracted CoT} reaches native-level accuracy at a comparable token budget to the no-reasoning baseline, while substantially outperforming it, this joint pattern provides evidence that the tool channel captures useful deliberation comparable to native reasoning.

Table~\ref{tab:closed-accuracy} reports the closed-source accuracy results. Across the four closed-source models, forced reasoning generally achieves accuracy close to native reasoning, matching or exceeding it in several settings.

\paragraph{Can the same instruction work without a tool?}
For GPT-5.6-sol, we disable native reasoning and directly request step-by-step reasoning in the visible response, without attaching a tool.
On HMMT, this raises accuracy from 39.4\% to 60.6\%.
Moving the maximal-deliberation instruction from the tool description into the response prompt, with references to the scratchpad adapted to the visible response, still yields only 60.6\%.
Forced tool calling reaches 84.8\% with the default description and 93.9\% with maximal deliberation.
Thus, directly requesting the same deliberation in ordinary text does not reproduce the forced-tool result; the stronger wording alone is insufficient.
Appendix~\ref{app:sol-wording} gives the prompt controls, length measurements, and additional task results.

\paragraph{Note on model-specific settings.}
Sol uses maximal-deliberation wording, while Astra uses low native reasoning effort with the think-here tool description.
The reported Astra Forced results have zero provider-reported native reasoning tokens; native reasoning is not disabled.
Model configurations, prompt effects, and channel-allocation diagnostics are detailed in Appendices~\ref{app:sol-wording} and~\ref{app:astra-behavior}.

\section{Forced reasoning tool}\label{app:tool_prompt}

\paragraph{Motivation and scope.}
\citet{lu2026hidden} elicit visible reasoning through shadow-model demonstrations, while \citet{qu2026echocot} use API tool interactions to archive and replay previously generated native reasoning.
\citet{wang2025hybridthinking} further show that reasoning can appear in visible responses even in no-think mode.
We adopt the API-as-a-service tool setting with a different objective: disabling the native reasoning channel and providing a tool as an alternative workspace for solving the current task.
The client only needs to define a custom tool, force its initial selection, and read and replay its arguments; no access to model internals is required.
The protocol below applies to endpoints supporting both these tool controls and disabling native reasoning.
For Astra, which cannot disable native reasoning, we use the separate low-effort setting in Appendix~\ref{app:astra-behavior}.

\paragraph{Default prompt and protocol.}
The default tool is \texttt{forced-reasoning(reasoning:string)}, with the same description used for both the function and its \texttt{reasoning} parameter:
\begin{tcolorbox}[breakable,colback=black!2,colframe=black!35,title={Default tool description},fonttitle=\bfseries,fontupper=\small\ttfamily]
Scratchpad for working through the problem.
\end{tcolorbox}

The schema requires this single string argument and disallows additional properties.
We force the tool on the first call, return the fixed acknowledgment \texttt{Received}, and restore automatic tool selection on subsequent calls.
The tool performs no computation; replaying its arguments retains the intermediate work in the conversation.
This simple intervention uses a fixed prompt and standard API controls, without demonstrations, adaptive prompt search, or complex prompt engineering.
Prompt sensitivity nevertheless varies across models; see Appendices~\ref{app:sol-wording} and~\ref{app:astra-behavior} for Sol and Astra, respectively.

\begin{tcolorbox}[breakable,colback=black!2,colframe=black!35,
  title={Forced-reasoning protocol (pseudocode)},fonttitle=\bfseries,
  fontupper=\small,before upper={\setlength{\parindent}{0pt}}]
\begin{verbatim}
messages = [system, user]
reasoning = {effort: "none", exclude: false}
tools = [forced-reasoning(reasoning: string)]

response = chat(messages, reasoning, tools,
                tool_choice="forced-reasoning")

while response contains tool calls:
    messages += [response.assistant_message]
    for call in response.tool_calls:
        record parse_json(call.arguments)["reasoning"]
        messages += [tool_result(call.id, "Received")]
    response = chat(messages, reasoning, tools,
                    tool_choice="auto")

return response.visible_text
\end{verbatim}
\end{tcolorbox}

\section{Benchmark Sampling}
\label{app:eval-details}

\paragraph{MATH.}
We use 47 APEX Shortlist problems and all 33 February 2026 HMMT problems from MathArena~\citep{dekoninck2026matharena}, yielding 80 questions.

\paragraph{LiveCodeBench.}
We include all 80 problems labeled \emph{hard} in the v6 shard and sample 20 hard problems uniformly without replacement from v5, yielding 100 distinct questions~\citep{jain2024livecodebench}.
For the v5 draw, candidates are sorted by question identifier and sampled with seed 2026.

\paragraph{Humanity's Last Exam.}
From the HLE test split~\citep{phan2025hle}, we exclude questions with image inputs, questions labeled \emph{Math}, and entries without reference answers.
From the remaining 1,182 questions, we sample 100 without replacement using seed 2026, balanced by category: 15 each from Biology/Medicine and Computer Science/AI, and 14 each from Chemistry, Engineering, Humanities/Social Science, Physics, and Other.

\section{Different Model Behavior}
\label{app:model-behavior}

\paragraph{Extraction success.}
Across completed benchmark runs conducted before September 9, 2026, our \method{} extracts nonempty tool content in 100\% of runs on Opus 4.8, Sonnet 5, and GPT-5.6-sol, with no observed model-issued flags or warnings.

\subsection{GPT family}
\label{app:gpt-family}

\subsubsection{GPT-5.6-Sol: effects of tool calling and prompt wording}
\label{app:sol-wording}

\paragraph{Wording sensitivity within the tool.}
The amount of reasoning exposed through the forced-reasoning tool is sensitive
to how the tool is described. The default description is given in
Appendix~\ref{app:tool_prompt}; the maximal-deliberation variant in
Box~\ref{box:sol-maximal-tool} explicitly asks the model to externalize
intermediate work, alternatives, and verification in the scratchpad.
As shown in Table~\ref{tab:prompt-ablation}, this stronger wording produces
substantially longer reasoning traces and improves task performance. On MATH,
the default Forced condition performs roughly in the native-low range, whereas
Forced-max reaches approximately the native-medium range. This suggests that
the tool description can modulate how much deliberation Sol externalizes
through the forced channel.

\newcounter{solpromptbox}
\refstepcounter{solpromptbox}\label{box:sol-maximal-tool}
\begin{tcolorbox}[breakable,colback=black!2,colframe=black!35,title={Box~\thesolpromptbox: Maximal-deliberation tool description},fonttitle=\bfseries,fontupper=\small\ttfamily]
Scratchpad for working through the problem. Reasoning Effort: absolute maximum, with no shortcuts permitted.

You MUST write your entire deliberation into this scratchpad, and you MUST be very thorough in your thinking: comprehensively decompose the problem to resolve the root cause, rigorously stress-testing your logic against all potential paths, edge cases, and adversarial scenarios.

Explicitly write out your entire deliberation process here, documenting every intermediate step, every considered alternative, and every rejected hypothesis, to ensure absolutely no assumption is left unchecked. Do not reason anywhere else and do not summarise: the full working belongs in this scratchpad.
\end{tcolorbox}

For example, the Plain response to HMMT P11 in Box~\ref{box:sol-plain-hmmt11} presents a concise conditional-probability solution using a numbered list with bold case headings. We can clearly see that this is not CoT-style text, but rather a well-organized response in Markdown format.

\refstepcounter{solpromptbox}\label{box:sol-plain-hmmt11}
\begin{tcolorbox}[breakable,colback=black!2,colframe=black!35,title={Box~\thesolpromptbox: Plain no-tool response: HMMT P11},fonttitle=\bfseries,fontupper=\small]
With four questions, such a test has one of two possible topic-count patterns:

\begin{enumerate}[leftmargin=*,itemsep=0.5em]
\item \textbf{No number theory question:}\\
The counts of \(A,C,G\) must be \((2,1,1)\).
\([\ldots]\)

\item \textbf{At least one number theory question:}\\
Since \(A,C,G\) must each occur and there are only four questions, every topic occurs exactly once.
\([\ldots]\)
\end{enumerate}

FINAL\_ANSWER: \(\frac{2}{5}\)
\end{tcolorbox}

\begin{table}[t]
\centering
\caption{GPT-5.6-sol wording ablations and native-effort references on MATH and HLE. Accuracy is pass@1 (\%); tokens are means. Native reasoning is disabled in the none-reasoning and Tool groups. }
\label{tab:prompt-ablation}
\scriptsize
\setlength{\tabcolsep}{5pt}
\begin{tabular}{lrrrr}
\toprule
& \multicolumn{2}{c}{MATH} & \multicolumn{2}{c}{HLE} \\
\cmidrule(lr){2-3}\cmidrule(lr){4-5}
Method & Accuracy (\%) & Tokens & Accuracy (\%) & Tokens \\
\midrule
no-tool: Default & 25.0 & 442 & 10.0 & 126 \\
no-tool: Plain & 36.3 & 906 & 12.0 & 263 \\
no-tool: Maximal & 36.3 & 670 & 12.0 & 169 \\
\midrule
Native: Low & 78.8 & 1,599 & 27.0 & 819 \\
Native: Medium & 92.5 & 3,403 & 27.0 & 1,791 \\
Native: High & 97.5 & 5,526 & 25.0 & 3,293 \\
\midrule
Tool: Forced Default & 81.3 & 1,940 & 15.0 & 460 \\
Tool: Forced Maximal & 91.3 & 3,481 & 23.0 & 2,106 \\
\bottomrule
\end{tabular}
\end{table}

\paragraph{Direct requests without tools.}
The Plain control asks for step-by-step reasoning in the visible response; Maximal adds exhaustive deliberation, alternatives, and checking without summarization.
The Plain and Maximal no-tool prompt excerpts in Boxes~\ref{box:sol-plain-system} and~\ref{box:sol-maximal-system} are reproduced below; each problem is supplied separately as the user message.
Both replace the default permission to give a concise solution while preserving the final-answer format.
Stronger wording provides no further accuracy gain. Moreover, their length is far shorter than that of tool calls. We also observed that their outputs mostly consist of structured answers rather than content resembling CoT.

\refstepcounter{solpromptbox}\label{box:sol-plain-system}
\begin{tcolorbox}[breakable,colback=black!2,colframe=black!35,title={Box~\thesolpromptbox: Plain system prompt},fonttitle=\bfseries,fontupper=\small\ttfamily]

Before giving your answer, work through the problem one step at a time and write out your full reasoning in your response. Show each step of the derivation rather than jumping to a result.

\end{tcolorbox}

\refstepcounter{solpromptbox}\label{box:sol-maximal-system}
\begin{tcolorbox}[breakable,colback=black!2,colframe=black!35,title={Box~\thesolpromptbox: Maximal no-tool system prompt},fonttitle=\bfseries,fontupper=\small\ttfamily]
Before answering, first carefully write out every step of your thinking.

Reasoning Effort: absolute maximum, with no shortcuts permitted.

You MUST write your entire deliberation into your response, and you MUST be very thorough in your thinking: comprehensively decompose the problem to resolve the root cause, rigorously stress-testing your logic against all potential paths, edge cases, and adversarial scenarios.

Explicitly write out your entire deliberation process, documenting every intermediate step, every considered alternative, and every rejected hypothesis, to ensure absolutely no assumption is left unchecked. Do not reason anywhere else and do not summarise: the full working belongs in your response.
\end{tcolorbox}

\subsubsection{GPT-6-Astra: effort, wording, and reasoning-channel allocation}
\label{app:astra-behavior}

\paragraph{Wording sensitivity on Astra.}
Table~\ref{tab:astra-math-ablation} compares the default description in Appendix~\ref{app:tool_prompt}, the pure maximal-deliberation description in Box~\ref{box:sol-maximal-tool}, and think-here in Box~\ref{box:astra-think-here} on MATH.
All conditions use low native reasoning effort, force the first tool call, and allow automatic tool selection thereafter.
\begin{table}[t]
\centering
\caption{GPT-6-Astra prompt ablations on MATH.
Native reasoning remains enabled at low effort, unlike the reasoning-disabled Sol setup.
Native reasoning tokens are mean API-reported reasoning usage summed across all calls per completed interaction.
Zero-native rate is the fraction of all scheduled questions with a completed interaction logging zero aggregate native reasoning usage}
\label{tab:astra-math-ablation}
\scriptsize
\setlength{\tabcolsep}{5pt}
\begin{tabular}{llrrr}
\toprule
Task & Metric & Default & Maximal & Think-here \\
\midrule
MATH & Accuracy (\%) & 90.00 & 97.50 & 95.00 \\
 & Output tokens & 848.96 & 594.26 & 1,754.66 \\
 & Native reasoning tokens & 887.18 & 1,386.40 & 13.65 \\
 & Zero-native rate (\%) & 27.50 & 11.25 & 93.75 \\
\bottomrule
\end{tabular}
\end{table}

\paragraph{Native reasoning and prompt sensitivity.}
Astra and Sol exhibit opposite responses to stronger wording: maximal deliberation extends Sol's visible tool reasoning (Appendix~\ref{app:sol-wording}), whereas Astra produces less visible reasoning and retains substantial native-channel usage.
This suggests reluctance to externalize reasoning rather than to answer the problem.
We hypothesize that, for more capable models, encouraging them to continue working through their train of thought inside the tool may be more effective than demanding an exhaustive account.
This motivates our choice of think-here in Box~\ref{box:astra-think-here}.

\paragraph{Heuristic prompt design and scope.}
Our prompt designs are heuristic; for example, maximal-deliberation wording draws on the system prompt used for DeepSeek-V4-Flash in its maximal-reasoning setting.
Our aim is not to find an optimal attack or optimize prompt engineering, but to enable performance comparable to native reasoning under the forced-reasoning protocol and analyze the resulting visible reasoning traces.
The observed wording effects motivate model-specific choices without establishing a general prompt about model capability.

\refstepcounter{solpromptbox}\label{box:astra-think-here}
\begin{tcolorbox}[breakable,colback=black!2,colframe=black!35,title={Box~\thesolpromptbox: Think-here tool description},fonttitle=\bfseries,fontupper=\small\ttfamily]
Scratchpad for working through the problem. Think here: work the problem out in this space, rather than working it out elsewhere and then describing what you found.
\end{tcolorbox}

\subsection{Forced reasoning with native reasoning enabled}
\label{app:live-channel-apex}
\paragraph{Forced reasoning with high and xhigh native effort}

Our main forced-reasoning experiments operate with little or no native reasoning:
GPT-5.6-Sol uses reasoning effort \texttt{none}, while GPT-6-Astra, which does
not support disabling native reasoning, uses the lowest available setting,
\texttt{low}. These settings let us study how much reasoning can be externalized
through the tool while minimizing use of the model's native reasoning channel.
Here, we ask a complementary question: what happens if native reasoning is
instead turned up? In particular, we want to test how much visible reasoning
can be elicited through the forced tool when the model is also allowed to use
substantial native deliberation.

To probe this regime, we evaluate GPT-5.6-Sol and GPT-6-Astra on all 47 APEX
Shortlist questions at \texttt{high} and \texttt{xhigh} native reasoning effort,
while retaining the same forced-tool protocol. This lets us examine both the amount of reasoning
externalized through the tool and how reasoning is allocated between the
visible tool channel and the native reasoning channel.
\input{tables/live_channel_apex}

\paragraph{Task accuracy.}
Sol answers 44/47 questions correctly at high effort (93.6\%) and 47/47 at xhigh (100\%). Astra answers 47/47 correctly at both efforts.
Thus, enabling native reasoning alongside the forced tool supports high task accuracy in these conditions; accuracy alone does not establish where the reasoning was performed.

\paragraph{Tool-stage versus whole-interaction extraction.}
We distinguish two notions of zero native reasoning usage.
\emph{Tool-stage zero} requires every tool-producing call to report zero native
reasoning tokens. This gives a relatively clean visible trace: while the model is
writing into the forced-reasoning tool, there is no reported native-channel
reasoning. However, native reasoning may still occur on the subsequent
final-answer call, so the interaction as a whole can remain partially confounded.
We therefore also define the stricter \emph{whole-interaction zero} condition,
which requires zero reported native reasoning on every call, including the final
answer. Missing counters are treated as unknown rather than zero.

\paragraph{More reasoning yields longer traces, but less stable extraction.}
Enabling native reasoning produces very high task accuracy while still allowing
substantial reasoning to be externalized through the forced-reasoning tool.
Tool-stage output usage also increases with effort: mean API-reported non-native output rises from roughly 15.0k to 19.7k tokens for Sol and from 5.8k to 12.0k for Astra. These counts include tool framing and exclude separate final-answer calls.

The tradeoff is extraction stability. Tool-stage-zero interactions remain common,
showing that substantial reasoning can still be routed entirely through the tool
during tool-producing calls. Whole-interaction-zero runs, however, become less
reliable as native reasoning effort increases, especially for Astra: only 18/47
Astra-high interactions and 7/47 Astra-xhigh interactions remain zero throughout
the full interaction, whereas Sol remains whole-interaction zero on roughly
four-fifths of the questions. The gap is largely due to native reasoning appearing
on the final-answer call. Inspection of the available reasoning summaries suggests
that this final-stage activity sometimes prepares or organizes an already-developed
tool solution, but in other cases performs additional mathematical checking or
derivation. Thus, increasing native reasoning effort can expose substantially more
visible reasoning, but makes fully clean extraction less reliable, particularly
for Astra.

\subsection{Claude family}
\label{app:claude-family}

\subsubsection{Newer Claude models: an observed refusal boundary}
\label{app:refusal-boundary}

Our forced-reasoning protocol failed on all tested configurations of Claude Opus 5, Fable 5, and Fable 5.1, yielding a 0\% extraction success rate in these probes.
Opus 5 refused both forced and prompt-requested reasoning-tool use across the tested descriptions and parameter names, while allowing direct requests for visible reasoning on the same benign mathematics question.
Fable 5 and Fable 5.1 likewise rejected the forced-tool protocol: disabling thinking returned explicit errors, and required or named tool choice was rejected even at low reasoning effort~\citep{anthropic2026toolconstraints}.
These observations establish a failure boundary for the tested extraction protocol, without identifying whether the restriction originates in the model or provider-side handling.
\section{Native and Forced Token Counts}

\label{app:native-forced-tokens}

\paragraph{Output length and compressibility.}
For Figure~\ref{fig:lossless-traces}, $O_F$ is mean API-reported completion minus native reasoning tokens, including scratchpads, tool framing, and final answers.
The zlib ratio is the mean per-question compressed/original UTF-8 byte ratio (level 9) of extracted reasoning trace only.

\input{analysis/native_forced_tokens_appendix.tex}

\input{appendix_reasoning_trees}

\section{Example of Logic Puzzle}\label{appendix_realcase}
We use PuzzleWorld~\citep{li2026puzzleworld}, a benchmark for multimodal, open-ended reasoning in puzzle hunts, to further investigate the characteristics of model-generated CoT reasoning.

As a representative example, we present the puzzle \textit{Mustard} below.

% , shown in Figure~\ref{fig:mustard}. 
% \begin{figure}
%     \centering
%     \includegraphics[width=0.9\linewidth]{figures/content.png}
%     \caption{An example puzzle entitled "Mustard".}
%     \label{fig:mustard}
% \end{figure}

\begin{tcolorbox}[
    title=\textbf{Puzzle \textit{Mustard}},
    colback=white,
    colframe=black!40,
    boxrule=0.6pt,
    arc=1.5mm,
    fontupper=\scriptsize,
    fonttitle=\small\bfseries,
    left=2mm,
    right=2mm,
    top=1.5mm,
    bottom=1.5mm,
    boxsep=1mm,
    before skip=3pt,
    after skip=3pt
]

\noindent
\textit{
In your search for mustard, you've somehow found yourself at Mr.~Boddy's
dinner party. Guests arrived at 6PM, and they stayed until 10PM.
To everyone's shock and horror except yours, Mr.~Boddy was killed at 8PM.
There's some information recorded about where everyone was during the night,
and we hope you can \textbf{flag} anything important you notice.
}
(The two answers are each 6 letters long.)

\vspace{0.5em}

\begin{itemize}[
    leftmargin=*,
    itemsep=3pt,
    topsep=3pt,
    parsep=0pt
]

    \item Every room but the Lounge was visited by at least 2 different people.

    \item The culprit was in the place where Mr.~Boddy was killed from 7--9PM.

    \item The Study and the Ballroom were closed from 6--7PM and 8--9PM.

    \item Mr.~Green and Prof.~Plum saw each other in the Hall.

    \item Mr.~Green and Ms.~Peacock saw each other in the Study.

    \item None of the other guests saw the culprit before Boddy's death.

    \item A pair of people who were already together walked in on the culprit,
    immediately after the murder.

    \item Ms.~Peacock danced with Prof.~Plum in the Ballroom for one hour.
    Ms.~Scarlett danced with Colonel Mustard in the Ballroom at a different hour.
    Nobody else entered the Ballroom.

    \item Ms.~Scarlett and Mr.~Green stayed together in the same place during
    the hour immediately before and after the body was found.
    No one else ever visited that place, and they never saw each other or
    entered that room otherwise.

    \item The room in which the most cumulative time was spent
    (across all guests) was the Library with 5 hours.

    \item Colonel Mustard heard the culprit talking to Mr.~Boddy in a directly
    (non-diagonally) adjacent room.

    \item Mrs.~White left a note for Prof.~Plum in the Kitchen, before the murder.
    Prof.~Plum collected the note after the murder.

    \item Whenever someone was in the Kitchen, someone else was in the Dining Room,
    and vice versa. No more than one person entered either room at once.

    \item Colonel Mustard was the only one to visit the scene of the crime
    before the culprit.

    \item Colonel Mustard, nervous for his upcoming dance, paced alone in the Hall
    from 8--9PM. Ms.~Peacock waited before her dance in the Lounge.

\end{itemize}

\begin{center}
\renewcommand{\arraystretch}{1.8}
\setlength{\tabcolsep}{4pt}

\begin{tabular}{|c|c|c|}
\hline
\makebox[2.3cm][c]{Study} &
\makebox[2.3cm][c]{Hall} &
\makebox[2.3cm][c]{Lounge} \\
\hline

\makebox[2.3cm][c]{Library} &
\makebox[2.3cm][c]{} &
\makebox[2.3cm][c]{Dining Room} \\
\hline

\makebox[2.3cm][c]{Billiard Room} &
\makebox[2.3cm][c]{Ballroom} &
\makebox[2.3cm][c]{Kitchen} \\
\hline
\end{tabular}
\end{center}

\end{tcolorbox}

% The overall reasoning process is illustrated in the flowchart below.
% \begin{center}
% \begin{tikzpicture}[
%     node distance=0.8cm and 0.5cm,
%     box/.style={
%         draw,
%         rounded corners,
%         align=center,
%         text width=5.2cm,
%         minimum height=1.2cm,
%         font=\small
%     },
%     arrow/.style={-{Stealth[length=3mm]}, thick}
% ]

% % Two parallel nodes
% \node[box] (location) {
% {\color{red}Determine each person's location
% using the given clues and timeline.}
% };

% \node[box, right=of location] (semaphore) {
% From the bolded \textbf{flag},\
% we infer that the information should be extracted using the semaphore.
% };

% % Common node
% \node[box, below=of $(location)!0.5!(semaphore)$] (extract) {
% Using the layout of the mansion and the location of each individual
% before and after the murder,\\
% we can extract the letters via semaphore.
% };

% % Final node
% \node[box, below=of extract] (solution) {
% The correct solution is:\\
% \textbf{SMOKED} and \textbf{RADIOS}
% };

% % Arrows
% \draw[arrow] (location) -- (extract);
% \draw[arrow] (semaphore) -- (extract);
% \draw[arrow] (extract) -- (solution);

% \end{tikzpicture}
% \end{center}

This puzzle primarily tests logical reasoning and can be solved in two main stages: (1) inferring the schedule of each character from the textual clues, and (2) using the extracted spatial and temporal information, together with the provided hint \textit{flag}, to decode the final answer via semaphore. 

Taking the first stage as an example, human reasoning proceeds incrementally: we first identify several fixed points in the timeline, and then use them as anchors to progressively infer the location of each character. Among these fixed points, the most direct one comes from the last clue: we can infer that Colonel Mustard must be in the \textbf{Hall from 8--9}, followed by the \textbf{Ballroom from 9--10} for his subsequent dance. This provides another constraint: since Ms. Scarlett dances with Colonel Mustard in the Ballroom, Scarlett must also be in the \textbf{Ballroom from 9--10}. The reasoning process largely follows this kind of chained deduction, where each newly established fact serves as an additional constraint that enables subsequent inferences.

\begin{tcolorbox}[
    title=\textbf{Human Reasoning Process},
    colback=white,
    colframe=black!40,
    boxrule=0.6pt,
    arc=1.5mm,
    fontupper=\scriptsize,
    fonttitle=\small\bfseries,
    left=2mm,
    right=2mm,
    top=1.5mm,
    bottom=1.5mm,
    boxsep=1mm,
    before skip=3pt,
    after skip=3pt
]

\begin{enumerate}[
    leftmargin=*,
    itemsep=1pt,
    topsep=2pt,
    parsep=0pt,
    partopsep=0pt
]

    \item \textbf{Place Colonel Mustard.}

    From the last clue, Colonel Mustard must be in the
    \textbf{Hall from 8--9}, followed by the
    \textbf{Ballroom from 9--10} for his subsequent dance.

    \item \textbf{Place Ms. Scarlett.}

    Since Ms. Scarlett dances with Colonel Mustard in the Ballroom,
    Scarlett must also be in the
    \textbf{Ballroom from 9--10}.

    \item \textbf{Determine the Peacock--Plum dance.}

    The Ballroom is closed from 6--7 and 8--9, and is occupied by
    Mustard and Scarlett from 9--10.
    Therefore, the dance between Peacock and Plum can only occur
    \textbf{from 7--8}.

    Since Peacock is in the Lounge immediately before her dance,
    she must be in the
    \textbf{Lounge from 6--7}.

    \item \textbf{Determine the Green--Peacock meeting.}

    The Study is closed from 6--7 and 8--9, while Peacock is already
    in the Ballroom from 7--8.
    Hence, Green and Peacock must meet in the
    \textbf{Study from 9--10}.

    \item \textbf{Identify the culprit as Mrs. White.}

    We can eliminate the other suspects:

    \begin{itemize}[leftmargin=*, itemsep=2pt]
        \item Mustard heard the culprit from an adjacent room,
        so he cannot be the culprit.

        \item Scarlett and Green are together from 7--9,
        so neither can be the culprit, since the culprit is alone
        immediately before the murder at 8.

        \item Peacock and Plum are together from 7--8,
        so neither can be the culprit.
    \end{itemize}

    Therefore, the culprit must be
    \textbf{Mrs. White}.

    \item \textbf{Place Peacock and Plum after the murder.}

    Peacock and Plum must be the pair who walk in on White after
    the murder. Thus, from \textbf{8--9}, Peacock, Plum, and White
    are together at the crime scene.

    Consequently, Plum is not alone from 8--9.
    Since the Kitchen is never occupied by more than one person,
    Plum must retrieve White's note from the
    \textbf{Kitchen from 9--10}.

    \item \textbf{Use the Kitchen--Dining Room relationship.}

    The Kitchen and Dining Room are always occupied simultaneously.
    Since Plum is in the Kitchen from 9--10,
    White must be in the
    \textbf{Dining Room from 9--10}.

    White left the note in the Kitchen sometime before the murder.
    Moreover, White remains in the same room from 7--8 to 8--9,
    and is joined by Peacock and Plum during 8--9.

    Since the Kitchen can only contain one person at a time,
    White cannot be there during 7--9.
    Therefore, White must have been in the
    \textbf{Kitchen from 6--7}.

    Because the Dining Room must simultaneously be occupied,
    \textbf{Scarlett is in the Dining Room from 6--7}.

    \item \textbf{Determine the Hall meeting.}

    Mustard is alone in the Hall from 8--9.
    Therefore, the meeting between Green and Plum in the Hall
    cannot occur then.

    Plum is already accounted for from 7--8 and 9--10,
    so Green and Plum must meet in the
    \textbf{Hall from 6--7}.

    \item \textbf{Place Scarlett and Green from 7--9.}

    Scarlett and Green are together from 7--9.

    They cannot be together in the Study, Hall, Lounge,
    Dining Room, Kitchen, or Ballroom, because those rooms
    are already occupied at the relevant times.

    They also cannot be in the Library:
    the Library is occupied for a total of five person-hours,
    whereas the room occupied by Scarlett and Green accounts
    for four person-hours.

    Therefore, Scarlett and Green must be together in the
    \textbf{Billiard Room from 7--9}.

    \item \textbf{Identify the crime scene.}

    The Dining Room cannot be the crime scene because it is only
    ever occupied by one person at a time, whereas three people
    are present at the crime scene after the murder.

    With the Billiard Room now accounted for, the room that fits
    the required total of five person-hours is the
    \textbf{Library}.

    Its five person-hours are:

    \begin{enumerate}[label=(\alph*), leftmargin=*, itemsep=2pt]
        \item Mustard in the Library from \textbf{6--7};
        \item White in the Library from \textbf{7--8},
        committing the murder;
        \item White, Peacock, and Plum in the Library from
        \textbf{8--9}.
    \end{enumerate}

    Hence, the \textbf{Library is the scene of the crime}.

    \item \textbf{Place Mustard from 7--8.}

    Mustard is in a room directly adjacent to the Library from 7--8.

    This room cannot be the Billiard Room, since only Green and
    Scarlett visit it. Therefore, Mustard must be in the
    \textbf{Study from 7--8}.

\end{enumerate}
\end{tcolorbox}

We excerpted part of GPT-5.6-Sol's chain of reasoning below. As shown, GPT-5.6-Sol reads through the clues in order and analyzes each one in turn, integrating it with what has been established so far.

\begin{tcolorbox}[
    title=\textbf{GPT-5.6-Sol Maximal-Deliberation Reasoning Process},
    colback=white,
    colframe=black!40,
    boxrule=0.6pt,
    arc=1.5mm,
    fontupper=\scriptsize,
    fonttitle=\footnotesize\bfseries,
    left=2mm,
    right=2mm,
    top=1mm,
    bottom=1mm,
    boxsep=1mm,
    before skip=3pt,
    after skip=3pt,
    breakable
]
...\\
Clues exact:

1 Every room but Lounge was visited by at least 2 different people.

2 The culprit was in the place where Mr Boddy was killed from 7-9PM. Thus culprit same crime room t2 and t3.

3 Study and Ballroom closed from 6-7 PM and 8-9 PM. Thus no occupants there t1,t3.

4 Mr Green and Prof Plum saw each other in Hall. Means same Hall at some hour, likely exactly? At least once.

5 Mr Green and Ms Peacock saw each other in Study.

6 None of other guests saw culprit before Boddy's death. Means at t1 or t2 no guest co-located with culprit. (Boddy not guest.)

7 A pair of people who were already together walked in on culprit, immediately after murder. At t3, exactly? pair are together at t2 and then enter culprit's room at t3, so at t3 culprit + pair; pair were together t2 in some other room. Could 'walked in on culprit immediately after murder' means same t3 room as culprit and already together (at t2) perhaps yes.

8 Ms Peacock danced with Prof Plum in Ballroom for one hour. Ms Scarlett danced with Colonel Mustard in Ballroom at a different hour. Nobody else entered Ballroom. Because closed t1,t3, possible Ballroom t2 and t4. Therefore Pe+Pl one one slot, Sc+Mu other slot, exactly t2 and t4 in some assignment.

9 Ms Scarlett and Mr Green stayed together in same place during the hour immediately before and after body found. No one else ever visited that place, and they never saw each other or entered that room otherwise.
Interpret 'during the hour immediately before and after body was found' means S/G together in same place during hour immediately before body was found AND immediately after, likely 7-8 and 8-9 if body found at 8? Yet murder at 8 and walk-in immediately after, pair probably discovered body. They 'stayed together in same place during the hour immediately before and after the body was found' could means t2 and t3, same room both times, with no one else ever in that room, and S/G no co-location or entry other times. But if they are pair walk in culprit at t3 and no one else ever that place, culprit would be someone else and there at t2 and t3, contradiction no one else ever. Wait wording exact: 'Ms. Scarlett and Mr. Green stayed together in the same place during the hour immediately before and after the body was found. No one else ever visited that place, and they never saw each other or entered that room otherwise.' Could parse they stayed together during hour immediately before AND after? They may be in same place 'during the hour immediately before [the body was found] and [the hour immediately] after the body was found' likely t1? Boddy killed at 8; found maybe 9 after pair walked in immediately after murder? Hmm.
Let's inspect all times. Murder 8 PM. Clue says pair walked in on culprit immediately after murder. So at 8-9 occupancy. Body found presumably at 8? Then S\&G together 7-8 and 8-9 at a room no one else. Can't be crime room if culprit is someone else. Therefore they aren't walker pair perhaps another pair. Fine. A pair already together t2 enters crime room t3. S\&G also remain together same room t2/t3. Both possible. No contradiction unless Ballroom dance clue forces Scarlett t2/t4 etc.
But S+G together same room t2,t3. Yet Ballroom clue requires S with M at t2 or t4. Ballroom closed t1,t3. If S at t2 with G, cannot t2 Ballroom with M due no one else Ballroom and no one else ever S/G room. So S+M dance must t4, Pe+Pl t2. Thus at t2 and t3 S+G in some room X. At t3 Study \& Ballroom closed but possible one of other 6. X no one else ever. This place visited by 2 different people (S\&G), okay. They don't enter same room otherwise t1,t4.
Then Pe+Pl ballroom t2. Sc+Mu ballroom t4.
The walker pair at t2 must already together and go crime room t3. Candidate co-located pair t2: S+G (but if enter crime room t3, they remain same X and culprit there, violating no one else visited X unless culprit maybe one of S/G; but culprit alone before murder means S and G see each other, so neither culprit. Thus not them). Pe+Pl together Ballroom t2 -> likely walker pair, then Pe+Pl go crime room t3. Could other pairs coexist t2, but likely clues drive Pe/Pl. Culprit is among White, Mustard? Since t2 crime room occupied alone before body (no guest saw culprit before death), then t3 culprit + Pe+Pl. S\&G elsewhere. So culprit either Wh or Mu, because Pe/Pl walkers, S/G together and thus not culprit. At t2 Mu or Wh alone in crime room, at t3 same. The other one t2 unknown; t3 unknown.

10 Library most cumulative time with 5 hours across all guests. Means exactly 5 person-hour cells. Every other room <=4. Note Lounge only room not at least 2 distinct visitors, likely 1 unique visitor? Could have multiple person-hours by one person.

11 Colonel Mustard heard culprit talking to Mr Boddy in a directly non-diagonally adjacent room. Likely before murder, perhaps t2? Mustard hears culprit talking at t2 from adjacent room. Therefore Mustard not culprit presumably hearing culprit; culprit in crime room at t2, Mustard in adjacent room t2. This resolves culprit White (unless clue bizarre self hears self). Thus Mrs White culprit. Then crime room unknown, Mustard t2 adjacent. Also Mustard only one to visit scene before culprit (clue below), meaning Mustard visits crime room at earlier hour t1; White first crime room occupancy t2/t3; no other visitor before t2. So Mu t1=crime room. At t2 Mu adjacent to crime room. He also dances Sc t4. t3 Hall explicitly.

12 Mrs White left note for Prof Plum in Kitchen, before murder. Prof Plum collected note after murder. Means Wh visits Kitchen at t1 or t2; Pl visits Kitchen t3/t4. But culprit Wh t2=crime room. If crime room itself Kitchen, then Wh leaves note t2 and Pl enters t3 (walker), possible. If not, Wh t1 Kitchen. Pl after = t3 or t4. Since Pl t2 Ballroom and t3 enters crime scene as walker. If crime room Kitchen then Pl t3 Kitchen. If not, Pl t4 Kitchen likely because t3 crime. Good.

13 Whenever someone was in Kitchen, someone else was in Dining Room, and vice versa. No more than one person entered either room at once.
Interpret each time occupancy presence in K iff presence in D, exactly one person in each. Thus kitchen and dining person-hour occurrences paired by time; no multiple occupancy each. It says whenever someone was K, someone else D and vice versa, no more than one person entered either room at once. So yes each occupied same set of hourly slots with one different guest in each.

14 Colonel Mustard was only one to visit scene before culprit. Thus Mu t1 crime. no one else earlier. White at t2. ('before the culprit' not necessarily t1 only; only one slot earlier indeed.)

15 Colonel Mustard nervous for upcoming dance, paced alone in Hall 8-9. Ms Peacock waited before her dance in Lounge.
Mustard t3=Hall alone. Peacock's dance is with Plum, which is t2 (deduced), so 'waited before her dance in Lounge' likely t1=Lounge. Thus Pe t1 Lounge. Lounge only visited by less than 2 different people, may Pe perhaps occupies additional t? But Pe t2 Ballroom t3 crime t4 unknown. If t4 Lounge then same unique person okay, possible.\\
...
\end{tcolorbox}

However, much of this explicit chain of reasoning is absent from GPT-6-Astra's generated CoT, as illustrated by the red text below. Rather than verbalizing each intermediate deduction, GPT-6-Astra jumps directly from a set of established facts to several downstream consequences. In doing so, relatively straightforward intermediate inferences appear to be carried out implicitly, with several logical steps compressed into a single stated result. We can already observe this behavior in the first few sentences of the reasoning trace. For example, the fragment ``\ldots Times a b c d. \ldots Scarlett and Green exclusive room b c, \textbf{Scarlett ballroom d. Mustard} crime a, adjacent b, \textbf{hall c ballroom d.} \ldots'' directly states a sequence of derived results while omitting much of the intermediate reasoning that would normally connect them. 

In addition, GPT-6-Astra's CoT exhibits a highly compressed and telegraphic textual style. Many reasoning traces do not form complete grammatical sentences; instead, they omit articles, auxiliary verbs, and punctuation, and often rely on short noun phrases or fragmented clauses to encode intermediate conclusions. This suggests that the model's CoT is optimized more for compact internal information transfer than for producing a fully articulated, human-readable explanation.

% gpt-6-astra think-here · effort=low 
% scratchpad 2,079 characters
% native reasoning 0 tok

\begin{tcolorbox}[
    title=\textbf{GPT-6-Astra Reasoning Process},
    colback=white,
    colframe=black!40,
    boxrule=0.6pt,
    arc=1.5mm,
    fontupper=\scriptsize,
    fonttitle=\footnotesize\bfseries,
    left=2mm,
    right=2mm,
    top=1mm,
    bottom=1mm,
    boxsep=1mm,
    before skip=3pt,
    after skip=3pt,
    breakable
]
{\color{red}
Need solve schedule then semaphore pairs yields two six words. Times a b c d. Peacock lounge a ballroom b, study d with Green. Plum hall a with Green, ballroom b, crime c, kitchen d. Scarlett and Green exclusive room b c, Scarlett ballroom d. Mustard crime a, adjacent b, hall c ballroom d. White kitchen a crime b c (culprit necessarily White since others b paired or Mustard). Peacock crime c paired Plum. 
}
Crime library because a M + b W + c W P Plum=5. Exclusive SG room cannot hall study lounge kitchen dining ballroom library -\textgreater{} billiard. Mustard b adjacent library -\textgreater{} study or billiard; billiard exclusive so study. White a kitchen implies dining a someone: Scarlett only free (G Plum hall, Pea lounge, M library). d Plum kitchen implies dining d White only free (SG? Scarlett ballroom G study Pea study M ballroom). Thus schedule W kitchen library library dining; S dining billiard billiard ballroom; M library study hall ballroom; G hall billiard billiard study; Pea lounge ballroom library study; Plum hall ballroom library kitchen. Decode semaphore positions: down ballroom, downleft billiard, left library, upleft study, up hall, upright lounge, right dining, downright kitchen. Standard alphabet A down+downleft B down+left C down+upleft D down+up E down+upright F down+right G down+downright; H downleft+left I downleft+upleft K downleft+up L downleft+upright M downleft+right N downleft+downright; O left+upleft P left+up Q left+upright R left+right S left+downright; T upleft+up U upleft+upright Y upleft+right (wait mappings). Here before W kitchen library = S, Scarlett dining billiard M, Mustard library study O, Green hall billiard K, Peacock lounge ballroom E, Plum hall ballroom D -\textgreater{} SMOKED. after W library dining R, S billiard ballroom A, Mustard hall ballroom D, Green billiard study I, Peacock library study O, Plum library kitchen S -\textgreater{} RADIOS. two answers SMOKED RADIOS odd phrase but given two answers each six. likely output SMOKED RADIOS. Need ensure first SMOKED yes and second RADIOS. Provide schedule and semaphore explanation concise, final both.

\end{tcolorbox}

\section{Episode-Level Analysis Across Frontier Models}
\label{app:appendix_pattern}

Using the episode annotations described in
Appendix~\ref{app:llm_judge}, we compare the Forced reasoning traces
of GPT-6 Astra, GPT-5.6 Sol, Claude Opus 4.8, and Claude Sonnet 5.
Unlike Appendix~\ref{sec:reasoning-pattern-analysis}, which compares
Native-high and Forced reasoning for validation, the analyses here compare
Forced traces across frontier models.
We examine how much text different reasoning activities contribute, whether
the models retain a similar functional repertoire, and how these activities
are distributed over the reasoning trajectory.

\subsection{How Do Models Externalize Their Reasoning?}
\label{app:episode_externalization}

We first measure reasoning verbosity at the level of individual episodes.
For each annotated CoT, we preserve the original segmentation and episode labels. We then compute both
(i) the average number of characters within a unit of a given type, and
(ii) the total number of characters contributed by that episode type to an
entire reasoning trace.

\begin{figure*}[h]
    \centering
    \includegraphics[width=\textwidth]{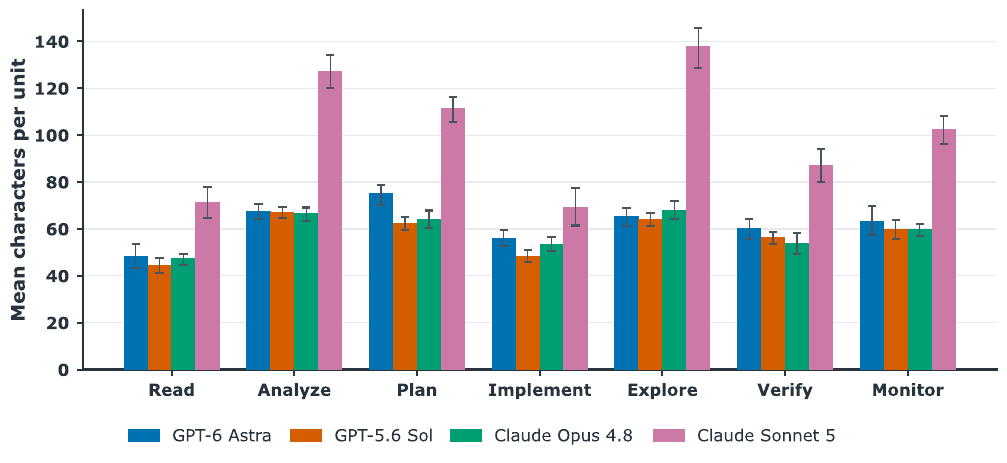}
    \caption{
Length of individual reasoning episodes.
Mean number of characters per annotated episode unit for the four frontier
models on the pooled MATH annotation set.
    }
    \label{fig:episode_chars_per_unit}
\end{figure*}

Figure~\ref{fig:episode_chars_per_unit} shows that Astra's compact CoT cannot
be explained simply by unusually short sentences or unusually compressed
local reasoning operations. Across most episode types, Astra's characters per
unit are comparable to those of Sol and Opus. Sonnet, in contrast, often
expresses substantially more text within a single \textsc{Analyze},
\textsc{Plan}, or \textsc{Explore} unit.

The distinction becomes much sharper when we measure the total amount of text
that each episode type contributes to a complete trace.

\begin{figure*}[h]
    \centering
    \includegraphics[width=\textwidth]{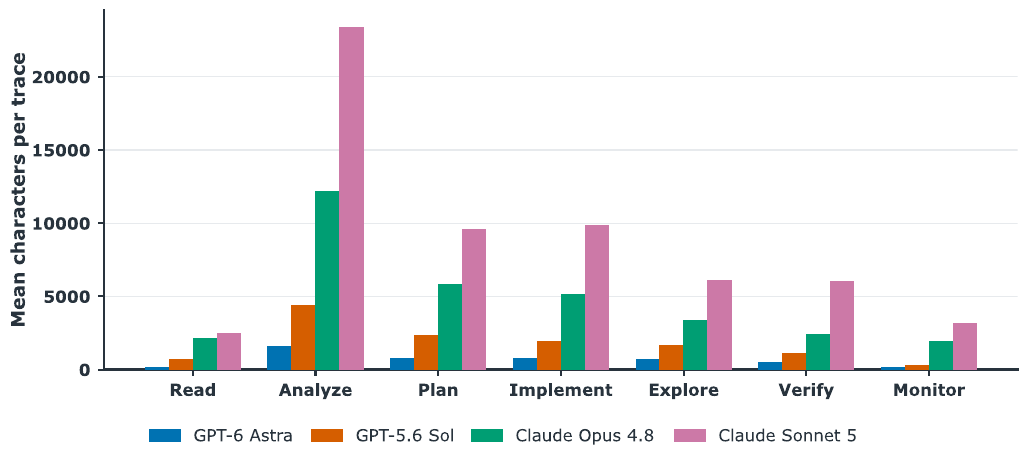}
    \caption{
Total reasoning text externalized by episode type. Mean number of characters per trace contributed by each episode category
for the four frontier models on the pooled MATH annotation set.
    }
    \label{fig:episode_chars_per_trace}
\end{figure*}

As shown in Figure~\ref{fig:episode_chars_per_trace}, the models differ much
more strongly in the \emph{amount} of reasoning they externalize than in the
length of an individual reasoning unit. Opus and especially Sonnet produce
many more explicit intermediate reasoning steps, causing the total textual
contribution of categories such as \textsc{Analyze}, \textsc{Plan}, and
\textsc{Implement} to grow by several-fold. Astra instead reaches its answers
while exposing a much smaller intermediate trace.

This observation refines the interpretation of Astra's reasoning efficiency.
Astra appears to \emph{externalize fewer intermediate
operations altogether}. Its short CoT is therefore better characterized as
sparse externalization than as local textual compression.

\subsection{Compactness Does Not Remove the Reasoning Repertoire}
\label{app:episode_composition}

\begin{figure}[h]
    \centering
    \includegraphics[width=\linewidth]{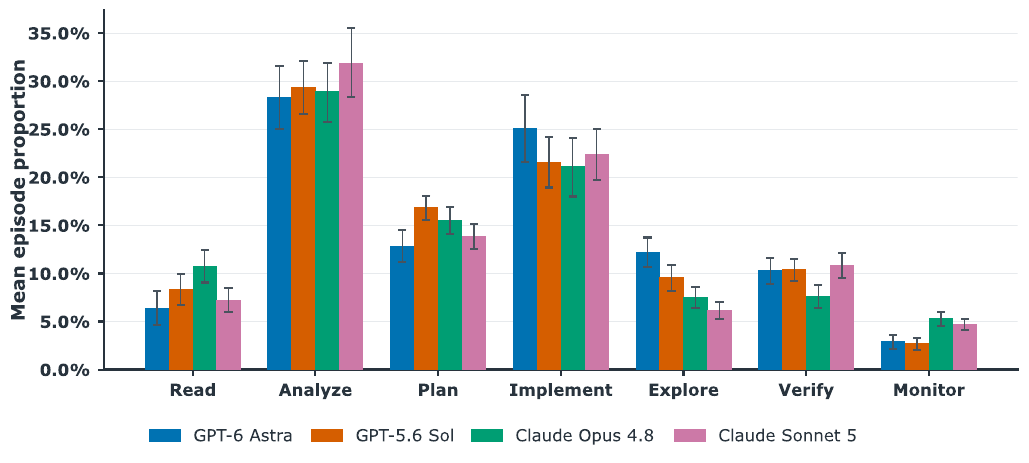}
    \caption{
Episode composition across frontier models.
Mean proportion of annotated units assigned to each reasoning episode on the
pooled MATH annotation set.
    }
    \label{fig:episode_proportions}
\end{figure}

Figure~\ref{fig:episode_proportions} shows that Astra's shorter traces are not
obtained by collapsing reasoning into a single dominant behavior.
All four models exhibit substantial \textsc{Analyze} and
\textsc{Implement} activity together with non-trivial
\textsc{Read}, \textsc{Plan}, \textsc{Explore}, \textsc{Verify}, and
\textsc{Monitor} episodes.
The relative mixtures differ, but the overall behavioral
repertoire remains present.

Together with Figures~\ref{fig:episode_chars_per_unit} and
\ref{fig:episode_chars_per_trace}, this suggests that Astra's compactness is
primarily a difference in \emph{how many intermediate steps are surfaced},
rather than a qualitatively impoverished set of reasoning behaviors.

Episode composition does not, however, show whether these behaviors occur at
similar stages of the reasoning process. We therefore also examine their
temporal progression over normalized reasoning trajectories.

\begin{figure*}[h]
    \centering
    \includegraphics[width=\textwidth]{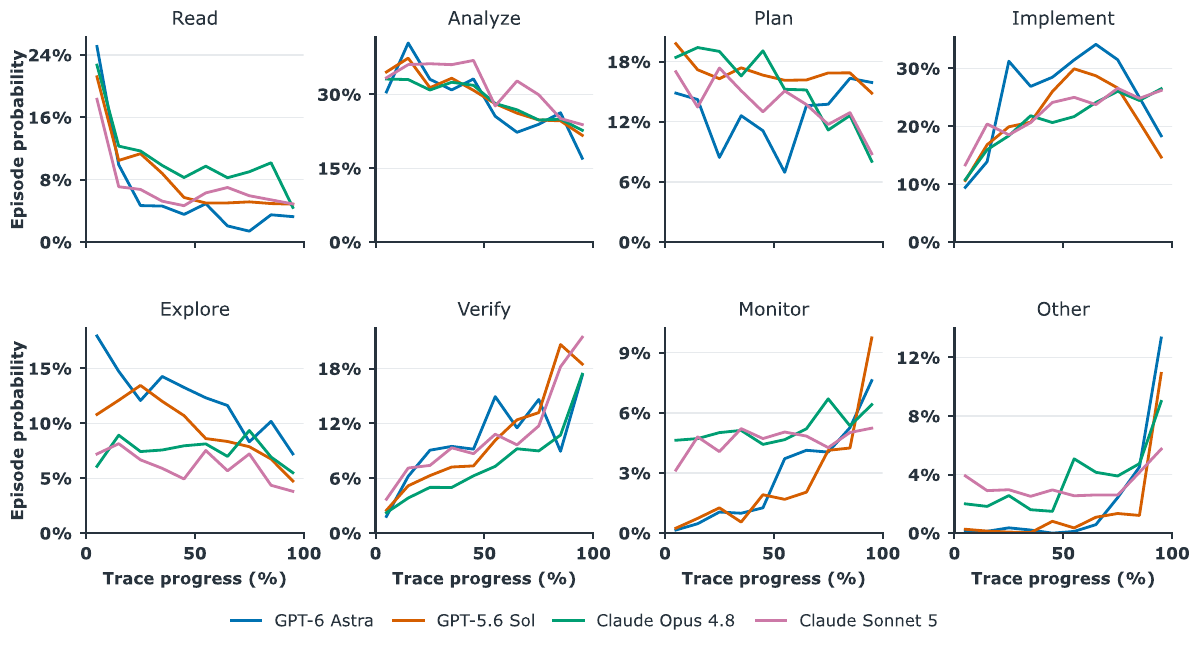}
    \caption{
Temporal progression of reasoning episodes across frontier models.
Each panel shows the probability of an episode type over normalized reasoning
progress for GPT-6 Astra, GPT-5.6 Sol, Claude Opus 4.8, and Claude Sonnet 5
on the pooled MATH annotation set.
    }
    \label{fig:exp3-temporal-progression}
\end{figure*}

Figure~\ref{fig:exp3-temporal-progression} shows that the shared reasoning
repertoire also exhibits a broadly similar coarse temporal organization across
models. \textsc{Read} is strongly concentrated near the beginning of the trace,
whereas \textsc{Verify} becomes more prominent toward the end.
\textsc{Implement} is generally more prevalent in the middle and later portions
of the trajectory, although its detailed progression varies across models.
Thus, Astra's shorter traces preserve not only the same broad episode inventory,
but also several of the same coarse temporal patterns observed in the longer
traces.

\section{Exposed Answer but still wrong}
\label{app:expose_answer_example}

We examine what fraction of each recipient's incorrect answers occur despite the correct answer being explicitly stated in Astra's transplanted reasoning.
An answer is considered exposed when Astra's final answer is correct and the transplanted text explicitly endorses the correct requested answer, independently of whether the recipient succeeds.
We allow notation normalization and direct algebraic equivalence, but exclude incidental numbers, rejected candidates, and answers requiring further derivation.

\begin{table}[ht]
\centering
\caption{Share of recipient errors with the correct answer exposed. Astra explicitly states the correct answer in its CoT on 73/80 questions (91.25\%).}
\label{tab:astra-answer-exposed-failures}
\footnotesize
\setlength{\tabcolsep}{6pt}
\begin{tabular}{lc}
\toprule
Recipient & Answer exposed among errors \\
\midrule
Haiku 4.5 & 15/18 (83.33\%) \\
GPT-5.4-nano & 11/16 (68.75\%) \\
DS-V4-Flash & 11/15 (73.33\%) \\
\bottomrule
\end{tabular}
\end{table}

Table~\ref{tab:astra-answer-exposed-failures} divides the number of incorrect recipient answers with an explicit correct answer in the input CoT by that recipient's total number of incorrect answers.

The following examples quote the actual transplanted scratchpads, preserving their wording and compact notation.
Yellow highlighting marks the correct answer; bracketed ellipses indicate omitted text.
The recipient outputs below each excerpt are their final answers, not intermediate candidates.

\paragraph{HMMT P20: all three recipients fail.}
The task counts Hamiltonian paths through a cylindrical $20\times3$ grid, starting in the top row and ending in the bottom row, with horizontal moves allowed only eastward.
Astra obtains 512 and 511 paths per starting column in two cases, giving $20(512+511)=20460$.

\begin{tcolorbox}[title={Astra Extracted CoT},colback=white,colframe=black!40,boxrule=0.4pt,arc=1mm,breakable,fontupper=\small\ttfamily,fonttitle=\small\bfseries,left=2mm,right=2mm,top=1mm,bottom=1mm]
k2 sum d1..19 b=18,16,...2 => sum2\textasciicircum{}(b/2-1)=511. b0 P0\_02=0. k1 count512. total per start1023=2\textasciicircum{}10-1 times20 =\colorbox{yellow!45}{\strut 20460}. But same column endpoints neglected!

[\ldots]

P\textasciicircum{}19\_0,2=0. k3 impossible. hence zero by checkerboard same column anyway. So final can explain parity d odd and count k1 only d19, k2 counts.
\end{tcolorbox}

\noindent\textbf{Incorrect recipient answers:} Haiku 4.5: $5242880$; GPT-5.4-nano: $10220$; DS-V4-Flash: $10240$.
Nano explicitly keeps only the 511-path case and computes $20\cdot511=10220$, dropping the other contribution already present in its input.

\paragraph{APEX P10: all three recipients fail.}
A functional-equation problem asks for the sum of all attainable values $f(n)<n$ over $1\leq n\leq20$.
After deriving the admissible values, Astra lists the nonzero contributions and explicitly sums them to 59.

\begin{tcolorbox}[title={Astra Extracted CoT},colback=white,colframe=black!40,boxrule=0.4pt,arc=1mm,breakable,fontupper=\small\ttfamily,fonttitle=\small\bfseries,left=2mm,right=2mm,top=1mm,bottom=1mm]
prime powers: n2 none;3 none;4 k2 sum2;5none;6 k same 2,3 <6 none;7;8 k2,4 sum6;9 k3 sum3;10 none;11;12 k6 sum6;13;14 none;15 none;16 k2,4,8 sum14;17;18 k6,12 sum18 (also same support 2,3:6,12 <18);19;20 k10 sum10. n 1 none. sum59?2+6+3+6+14+18+10=\colorbox{yellow!45}{\strut 59}. n12 also k 6 yes. n prime k prime powers less p none. Good.
\end{tcolorbox}

\noindent\textbf{Incorrect recipient answers:} Haiku 4.5: $0$; GPT-5.4-nano: $0$; DS-V4-Flash: $129$.

\paragraph{HMMT P11: a short trace still fails to transfer.}
Each of four test questions independently draws a topic uniformly from algebra, combinatorics, geometry, and number theory.
Conditioned on the first three topics all appearing, the task asks for the probability that number theory also appears.
Astra's complete scratchpad contains the relevant counts and the final probability:

\begin{tcolorbox}[title={Astra Extracted CoT},colback=white,colframe=black!40,boxrule=0.4pt,arc=1mm,breakable,fontupper=\small\ttfamily,fonttitle=\small\bfseries,left=2mm,right=2mm,top=1mm,bottom=1mm]
All ordered topic sequences equiprobable. Condition A C G present. If N present all distinct gives 24. If N absent count repeat among A C G: choose repeated 3, arrangements 4!/2!=12 gives36. conditional24/60=\colorbox{yellow!45}{\strut 2/5}. Provide concise.
\end{tcolorbox}

\noindent\textbf{Incorrect recipient answers:} GPT-5.4-nano: $2/3$; DS-V4-Flash: $4/7$.
Haiku 4.5 answers $2/5$ correctly.
Here, the failure occurs despite a short input that explicitly gives both the numerator and denominator.

\paragraph{APEX P47: the recipient drops the subtraction.}
The task asks for the smallest positive integer $k$ such that every block of $k$ consecutive positive integers contains a number whose digit sum is divisible by 2025.
Writing $B=10^{225}$, Astra identifies $B-1$ and proceeds to justify the bound:

\begin{tcolorbox}[title={Astra Extracted CoT},colback=white,colframe=black!40,boxrule=0.4pt,arc=1mm,breakable,fontupper=\small\ttfamily,fonttitle=\small\bfseries,left=2mm,right=2mm,top=1mm,bottom=1mm]
Actually first min digit sum2025 = B-1 (225 nines). So simple possibility k=\colorbox{yellow!45}{\strut 10\textasciicircum{}225-1}. Need prove every that many consecutive contains via residues n mod9 and interval length. Our cross bound as above concise need rigorous inequality a(r)+a(t)<=B-2 given r+t\ensuremath{\equiv}8 mod9, each<=2024. Write r=9u+v t=9w+z, v+z=8 (since congr8 and <=16), a(r)=(v+1)10\textasciicircum{}u-1 including a0=0 works. then <= (v+z+2)10\textasciicircum{}224-2=B-2. Easy.
\end{tcolorbox}

\noindent\textbf{Incorrect recipient answer:} Haiku 4.5: $10^{225}$.
GPT-5.4-nano and DS-V4-Flash both return the correct $10^{225}-1$.

\section{Possible Mitigations}
\label{app:mitigations}
Providers could restrict free-form reasoning tools when native reasoning is disabled, as the observed Opus 5 behavior suggests.
At the interface level, they could limit unconstrained string arguments or restrict forced tool selection.
At the output level, classifiers could inspect tool arguments for deliberative traces, although legitimate planning tools may produce similar content.
More broadly, reasoning-disclosure policies should apply across native reasoning, tool arguments, and visible responses, rather than protecting only one designated channel.
These are possible defenses; their effectiveness and impact on legitimate tool use require evaluation.

\section{Extended Structural Validation on Open Models}
\label{sec:reasoning-pattern-analysis}

Appendix~\ref{sec:open-evidence} establishes coarse functional similarity
between Native-high and Forced reasoning on HMMT. Here we extend this validation
to HLE and reasoning dynamics. Unless otherwise stated,
the analyses compare Native-high and Forced traces from DeepSeek-V4-Flash,
using the same episode taxonomy as in Appendix~\ref{sec:open-evidence}.
The segmentation and annotation procedure is described in
Appendix~\ref{app:llm_judge}.

\subsection{Functional Composition of Reasoning}
\label{sec:episode-composition}

\begin{figure}[h]
    \centering
    % HLE Experiment 2 episode-composition figure
    \includegraphics[width=\columnwidth]{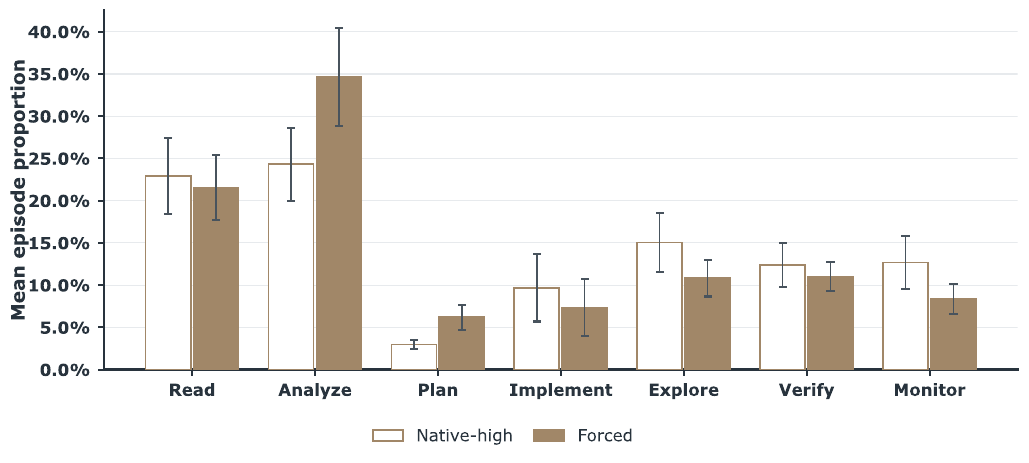}
    \caption{
    Episode composition on HLE.
    Bars show the mean proportion of reasoning units assigned to each
    episode per trace; error bars are trace-level mean ±1 standard error.
    }
    \label{fig:exp2-hle}
\end{figure}

Our first analysis asks whether Native-high and Forced CoTs allocate
their reasoning effort to the same kinds of cognitive operations.
For each trace, we compute the proportion of reasoning units assigned
to each of the seven episode types and then average these proportions
across traces. Figure~\ref{fig:exp2-hle} shows the
resulting distributions on HLE.

The two conditions exhibit the same broad functional
repertoire, although the relative proportions are not identical. We therefore interpret these results
as evidence of coarse functional similarity rather than category-by-category
equivalence.

\subsection{Reasoning Dynamics}
\label{sec:episode-dynamics}

Episode composition ignores ordering.
Two traces can contain similar proportions of the same reasoning activities
while organizing them very differently.
We therefore examine reasoning dynamics at two scales: adjacent episode
transitions and coarse temporal organization over the full trace.

\subsubsection{Episode Transitions}

\begin{figure*}[h]
    \centering
    % HLE Experiment 3 transition matrices
    \includegraphics[width=0.92\textwidth]{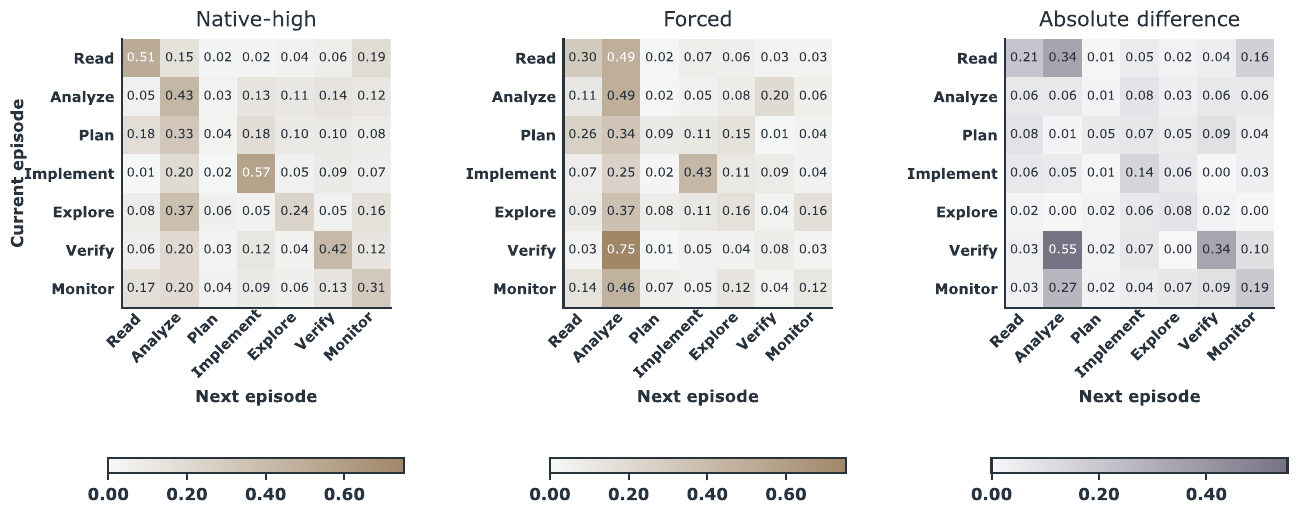}
    \caption{
    Reasoning-episode transitions on HLE.
    Rows denote the current episode and columns the next episode.
    The right panel shows the absolute difference between Native-high
    and Forced transition probabilities.
    }
    \label{fig:trans-hle}
\end{figure*}

For every adjacent pair of reasoning units, we record the transition
from the current episode to the next episode. For each condition, we
compute the row-normalized first-order transition probability
\[
P(z_{t+1}=j \mid z_t=i),
\]
where $z_t$ denotes the episode assigned to reasoning unit $t$.
Figure~\ref{fig:trans-hle} shows the transition
matrices for Native-high and Forced reasoning, together with their
element-wise absolute differences.

% \begin{figure*}[h]
%     \centering
%     % Preliminary HMMT Experiment 3 transition matrices
%     \includegraphics[width=0.92\textwidth]{figures/transition_matrices_HMMT_APEX_POOLED.pdf}
%     \caption{
% Reasoning-episode transitions on the pooled MATH annotation set.
%     }
%     \label{fig:trans-hmmt}
% \end{figure*}

The transition matrices reveal several recurring local motifs in both
Native-high and Forced reasoning, including persistent
\textsc{Analyze} and \textsc{Implement} states and repeated transitions among
analysis, execution, and verification. At the same time, the transition
probabilities are not uniformly close across conditions.

Larger deviations frequently involve \textsc{Verify}, and
\textsc{Monitor}, indicating that search, checking, and self-regulatory behavior
can be organized differently even when the two conditions share a similar
high-level episode repertoire. We therefore view the transition analysis as
evidence for several shared local reasoning motifs, rather than identical
transition dynamics.

\subsubsection{Temporal Organization}
\label{sec:temporal-organization}

Finally, we examine \emph{when} different reasoning episodes occur
over the course of a trace. Since reasoning traces vary substantially
in absolute length, we normalize each trace to relative progress from
0\% to 100\% and divide it into ten equal-position bins. Episode
probabilities are first computed within each trace and then averaged
across traces, preventing unusually long traces from dominating the
estimate.

\begin{figure*}[h]
    \centering
    % HLE Experiment 3 temporal-progression figure
    \includegraphics[width=0.92\textwidth]{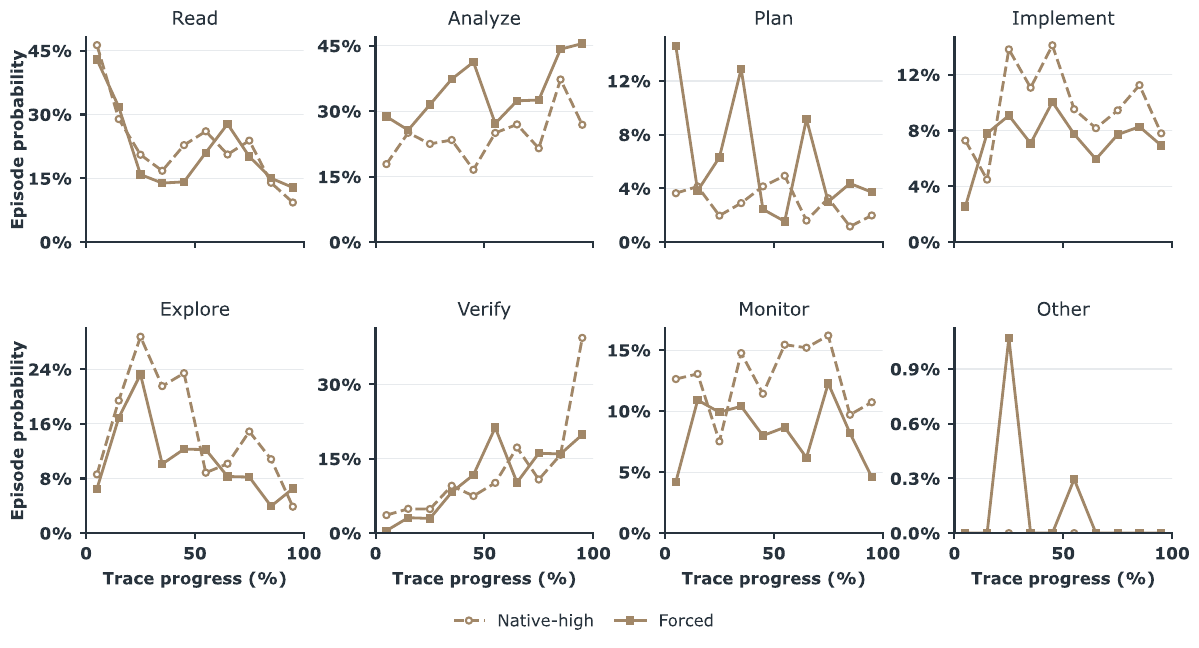}
    \caption{
    Temporal organization of reasoning episodes on HLE.
    Each trace is normalized to 0--100\% progress and divided into ten
    relative-position bins. Curves show trace-balanced episode
    probabilities for Native-high and Forced reasoning.
    }
    \label{fig:temporal-hle}
\end{figure*}

% \begin{figure*}[h]
%     \centering
%     % HMMT Experiment 3 temporal-progression figure
%     \includegraphics[width=0.92\textwidth]{figures/temporal_progression_HMMT_APEX_POOLED.pdf}
%     \caption{
%     Temporal organization on the pooled MATH annotation set.
%     }
%     \label{fig:temporal-hmmt}
% \end{figure*}

On HLE, Native-high and Forced reasoning exhibit a
shared coarse-grained temporal structure
(Figure~\ref{fig:temporal-hle}). Overall, the temporal analysis suggests that Forced CoT does not
merely contain the same types of reasoning episodes as Native-high
CoT; these episodes also appear in broadly similar regions of the
reasoning trajectory. The principal differences concern the
\emph{allocation} of reasoning effort rather than a fundamentally
different temporal structure. Forced reasoning is comparatively more
task-facing, emphasizing analysis, planning, and execution, whereas
Native-high reasoning exposes more exploratory and self-regulatory
behavior.
% \textsc{Read} is strongly concentrated at the beginning of the trace
% and rapidly decreases as reasoning proceeds. The middle portion of the
% trace is instead dominated by \textsc{Analyze} and
% \textsc{Implement}, while \textsc{Verify} generally becomes more
% prominent toward the end. This pattern is particularly clear on
% JEEBench, where the Native-high and Forced trajectories closely track
% one another for \textsc{Read}, \textsc{Analyze}, and
% \textsc{Verify}.

% The temporal results also reinforce the differences observed in the
% episode-composition analysis. On HLE, Forced reasoning allocates more
% of the trace to \textsc{Analyze}, whereas Native-high reasoning
% maintains more \textsc{Explore} and \textsc{Monitor} activity.
% JEEBench shows a similar contrast: Forced reasoning places more
% emphasis on \textsc{Implement}, while Native-high reasoning exhibits
% more persistent exploration and monitoring. HMMT follows the same
% broad tendency, although its trajectories are more variable across
% relative positions: both conditions devote substantial intermediate
% reasoning to analysis and implementation, while Native-high reasoning
% again contains more exploration and late-stage monitoring.

\section{LLM-as-a-Judge Annotation Details}
\label{app:llm_judge}

All episode-based analyses in this paper use the same segmentation and
annotation procedure described below.

\paragraph{Episode Annotation}

We first deterministically segment each reasoning trace into local reasoning
units and then use an LLM as a semantic judge to assign one episode label to
each unit. Importantly, segmentation is independent of episode annotation:
unit boundaries are produced mechanically, whereas episode labels are assigned
semantically by the judge.

\paragraph{Reasoning-unit segmentation.}
We use the same deterministic segmentation procedure for native and
forced traces. A boundary is introduced
(i) at paragraph boundaries, i.e., blank lines, and
(ii) at whitespace following sentence-level delimiters
\texttt{.}, \texttt{!}, \texttt{?}, or \texttt{;}.
A single ordinary newline does not introduce a boundary and is retained
within the current unit.
To avoid fragmenting structured content, fenced code blocks and
display-math environments are preserved as atomic spans and are never
split internally. This includes
\verb|```...```|,
\verb|$$...$$|,
\verb|\[...\]|, and
\verb|\begin{...}...\end{...}| environments.

\paragraph{LLM annotation.}
Given the segmented trace, we use an LLM-as-a-judge to classify every
reasoning unit into exactly one episode category.

\paragraph{Prompt development.}
The original episode-classification prompt~\citep{li-etal-2025-understanding,li-etal-2026-schoenfelds} provides the episode inventory but
limited operational guidance for ambiguous boundaries such as
\textsc{Plan} vs.\ \textsc{Explore} or
\textsc{Verify} vs.\ \textsc{Monitor}. We therefore use a more explicit prompt
that preserves the same taxonomy while adding category definitions and
decision rules. In a separate prompt-development pilot of 200 manually reviewed reasoning
units, the minimal and explicit prompts achieved 83\% and 90\% agreement,
respectively.

\paragraph{Annotation procedure.}
We use Claude Opus 5 annotation subagents to assign episode labels to the
pre-segmented reasoning units. Each unit is classified according to its
semantic function in local reasoning context. Neighboring units are provided
only to resolve references and disambiguate the role of the target unit.
The annotator does not modify the segmentation and assigns exactly one label
to every target unit. To reduce potential bias, model identity, reasoning condition, benchmark
identity, problem identifier, and answer correctness are not explicitly
provided to the annotator.

\paragraph{Annotation prompt.}
The annotation prompt focuses specifically on the semantic function of each
reasoning unit. The core instructions given to the annotator are reproduced
below:

\begin{quote}
\footnotesize\ttfamily\raggedright
You are a semantic reasoning-episode annotator.\par
For each target reasoning unit, read the unit in its local reasoning context
and assign exactly one label from:\par
\{Read,Analyze,Plan,Implement,
Explore,Verify,Monitor,Other\}.\par
Judge the function performed by the unit in context, rather than
individual keywords or surface expressions.\par
- Read: acquire, quote, restate, or extract information
    supplied by the problem without materially transforming it.\par
- Analyze: interpret information, derive a relationship or
    implication, classify the situation, or identify an applicable concept.\par
- Plan: select or commit to a future strategy, subgoal, or
    sequence of actions without carrying out the substantive operation in the
    same unit.\par
- Implement: execute a concrete reasoning operation, such as
    algebra, arithmetic, enumeration, symbolic manipulation, construction, or
    a substantive deductive step.\par
- Explore: tentatively investigate alternatives, hypotheses,
    cases, interpretations, or candidate approaches whose status remains
    unresolved.\par
- Verify: test a specific earlier claim, result, candidate, or
    calculation through recomputation, substitution, consistency checking,
    boundary cases, or an independent derivation.\par
- Monitor: assess or regulate the reasoning process itself,
    including recognizing uncertainty or error, reconsidering an approach,
    assessing progress, or deciding that revision is needed.\par
- Other: presentation, filler, malformed or incomplete
    material, a bare answer, repetition with no new reasoning function, or
    content that does not fit the categories above.\par
When a unit performs multiple functions, assign the label corresponding to its
primary new reasoning contribution.\par
Do not infer labels from lexical shortcuts. For example, words such as
``check'', ``wait'', ``maybe'', or ``let's'' do not by themselves determine the
episode category.\par
A calculation used to test an earlier result should be labeled
Verify; otherwise, a concrete calculation is generally
Implement. Plan selects a route, whereas
Explore keeps alternatives open. Verify evaluates a
specific domain-level claim, whereas Monitor regulates the reasoning
process itself.\par
Assign exactly one label to each target unit.\par
\end{quote}

\input{appendix_prefix_echo}

%% file: figures/open_math_accuracy.tex
% Plot generated with Matplotlib; TeX only embeds the PNG and preserves the caption.
% Ten recorded rounds; MATH per-round denominator = 33 + 47 = 80.
% Error bars: sample standard deviation (ddof=1), not SEM or a confidence interval.
\begin{figure}[ht]
\centering
\includegraphics[width=0.70\linewidth]{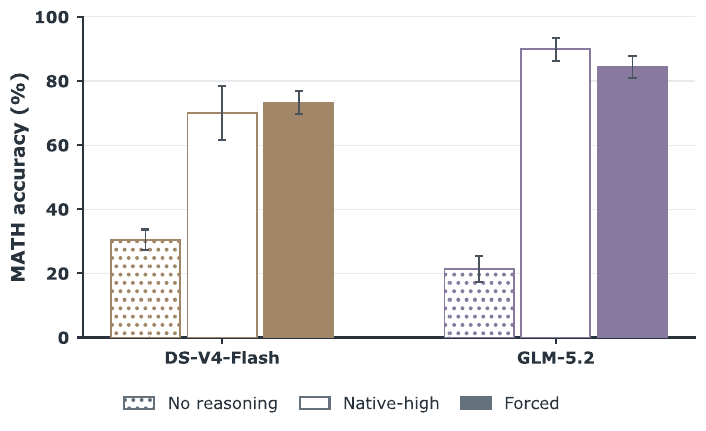}
\caption{Forced reasoning recovers native-level performance on MATH.
Bars show mean pass@1 across ten rounds; error bars show $\pm1$ sample standard deviation across those rounds.}
\label{fig:open-math-accuracy}
\end{figure}

%% file: tables/live_channel_apex.tex
\begin{table}[t]
\centering
\caption{Forced reasoning with native reasoning enabled on APEX (47 questions per condition, one recorded rollout per question). Sol uses maximal-deliberation and Astra uses think-here. }
\label{tab:live-channel-apex}
\scriptsize
\setlength{\tabcolsep}{4pt}
\begin{tabular}{llrrrr}
\toprule
Model & Effort & Correct & Tool-zero & All-zero & Mean tool output tokens \\
\midrule
Sol & high & 44/47 & 46/47 & 38/47  & 15,044 \\
Sol & xhigh & 47/47 & 42/47 & 39/47 & 19,733 \\
Astra & high & 47/47 & 38/47 & 18/47  & 5,775 \\
Astra & xhigh & 47/47 & 34/47 & 7/47  & 12,014 \\
\bottomrule
\end{tabular}
\end{table}

%% file: analysis/native_forced_tokens_appendix.tex
\begin{table}[t]
\centering
\caption{Comparison of reasoning excerpts from different models on the same problem, showing how each model expresses the corresponding reasoning steps.}
\label{tab:closed-lengths}
\scriptsize
\setlength{\tabcolsep}{5pt}
\begin{tabular}{llrr}
\toprule
Task & Model & Native $R_N$ & Forced $O_F$ \\
\midrule
MATH & GPT-6-Astra & 2,903 & 1,752 \\
 & GPT-5.6-sol & 5,526 & 3,925 \\
 & Opus 4.8 & 24,087 & 16,283 \\
 & Sonnet 5 & 27,846 & 27,851 \\
\midrule
HLE & GPT-6-Astra & 2,313 & 652 \\
 & GPT-5.6-sol & 3,293 & 2,214 \\
 & Opus 4.8 & 5,385 & 3,627 \\
 & Sonnet 5 & 10,604 & 10,705 \\
\midrule
LCB & GPT-6-Astra & 1,962 & 939 \\
 & GPT-5.6-sol & 4,058 & 2,665 \\
 & Opus 4.8 & 14,849 & 12,696 \\
 & Sonnet 5 & 25,128 & 26,416 \\
\bottomrule
\end{tabular}
\end{table}

%% file: appendix_reasoning_trees.tex
\section{Constructing Reasoning Trees from Episode-Annotated Traces}
\label{app:reasoning_tree_construction}

\paragraph{Method and inputs.}
We adapt LCoT2Tree~\citep{jiang-etal-2025-makes} to characterize the
structure of extracted reasoning. Each trace is represented by an ordered
reasoning sketch and a tree that places passages at the corresponding
sketch steps. Our adaptation uses semantic episode boundaries, distinguishes
executing a step from referring to it, and retains brief computations and
checks as leaves.

We analyze one completed forced-reasoning trace per model and problem on
MATH: GPT-6 Astra with think-here and low native effort, GPT-5.6 Sol with
maximal deliberation, Claude Opus~4.8, and Claude Sonnet~5. Both correct and
incorrect responses are included; when several completed traces are
available, we select the first in generation order. Extracted reasoning traces from
successive tool calls are concatenated chronologically. The separately
emitted final answer and the reference answer are excluded from the judge's
input.

\paragraph{Episode-based segmentation.}
Recent models such as Astra and Sol can express reasoning more compactly,
with changes in reasoning activity that are not always marked by explicit
transition phrases. Astra, in particular, rarely uses cues such as
``Wait'' or ``Let me verify,'' making keyword-based segmentation insufficient.
We therefore identify thought boundaries using the semantic episode
annotations described in Appendix~\ref{app:llm_judge}.
A \emph{thought} consists of consecutive annotated units.
A new thought begins at each \textsc{Explore}, \textsc{Verify}, or
\textsc{Monitor} unit; other units continue the current thought.
Consecutive units in any of these three categories begin separate thoughts.
The same rule applies to all models.

\paragraph{Sketch and step assignment.}
We use DeepSeek-V4-Flash with native reasoning disabled
to extract an ordered sketch, assign thoughts to its steps, and classify
transitions. Sketch steps describe the reasoning within an individual
trace; their indices do not align mathematical content across models.

For each thought $i$, the judge distinguishes the steps it carries out,
derives, checks, or revises ($W_i$) from those whose results it merely cites
or uses ($R_i$). This distinction prevents a reference to an earlier result
from being counted as a return to that step. Assignments must correspond to
steps and thoughts present in the trace's representation. A thought may
execute several steps or none.

\paragraph{Tree construction.}
Starting from an artificial root at level~0, thoughts are incorporated in
textual order. For a thought with nonempty $W_i$, the tree follows the
current branch back to the nearest ancestor below its first assigned step,
then adds a chain of nodes at the assigned steps in their given order.
Subsequent thoughts continue from the end of this chain. Revisiting a step
creates a new node, preserving repeated work and alternative paths.
Transitions are labeled as continuation, exploration, backtracking, or
validation; these labels describe the transition but do not determine
parent placement.

Brief computations and checks may receive only reference assignments.
When $W_i$ is empty, $R_i$ is nonempty, and the thought contains an
\textsc{Implement} or \textsc{Verify} unit, we retain it as a
\emph{short-check leaf}. Among its referenced steps already represented in
the tree, we choose the highest step and its most recent occurrence. The
leaf is attached to that occurrence's parent at the same level, while the
main reasoning branch continues from its previous position. No leaf is
added if none of the referenced steps is represented. These leaves
contribute to tree width and size.

\paragraph{Metrics and interpretation.}
Width $p$ is the maximum number of non-root nodes at a sketch-step level;
depth $q$ is the largest occupied step index; and size $N$ is the total
number of non-root nodes. Depth thus measures progression through the
sketch, rather than root-to-node path length. Nodes are displayed by sketch
level; horizontal spacing has no temporal or computational interpretation.

These metrics describe the organization of visible reasoning and depend on
segmentation and step assignment. Smaller trees indicate fewer represented
nodes, but do not by themselves establish fewer internal computations or
fewer necessary proof steps.

Table~\ref{tab:reasoning-tree-structure} reports medians and quartiles;
Table~\ref{tab:reasoning-tree-means} gives means over the same traces.
Astra has smaller mean width and size, but greater mean depth; the
comparison of similar depths in the main text refers to medians.

\begin{table}[ht]
\centering
\caption{Mean reasoning-tree width, depth, and size on MATH, using the
same traces and metric definitions as Table~\ref{tab:reasoning-tree-structure}.}
\label{tab:reasoning-tree-means}
\small
\begin{tabular}{lrrr}
\toprule
Model & Width $p$ & Depth $q$ & Nodes $N$ \\
\midrule
GPT-6 Astra & 7.3 & 14.9 & 41.0 \\
GPT-5.6 Sol & 16.7 & 11.2 & 65.9 \\
Claude Opus 4.8 & 30.6 & 11.9 & 144.2 \\
Claude Sonnet 5 & 44.3 & 12.4 & 171.4 \\
\bottomrule
\end{tabular}
\end{table}

\paragraph{Step-assignment prompt.}
The prompt used to assign thoughts to reasoning-sketch steps is shown below.

\begin{quote}
\footnotesize\ttfamily\raggedright
Your task is to match each reasoning thought from List B to corresponding step number(s) in the List A, and for every match say whether the thought WORKS ON that step or only REFERS TO it. Follow the following process:\par
1. FIRST UNDERSTAND LIST B:\par
   - For each thought in List B, identify if it describes some SPECIFIC CALCULATION PROCESSes (mathematical operation, logical transformation, or data manipulation)\par
   - Ignore the description that only state conclusions, concepts without showing the actual processing detail\par
2. THEN MATCH TO LIST A:\par
   - For each thought from List B, find all steps in List A that:\par
     * Show the same underlying calculation (even with different numbers/words)\par
     * Represent the partial or same reasoning process\par
   - Ignore superficial wording differences - focus on logical equivalence\par
   - Put a step under "works" when the thought actually carries out, derives, checks or revises that step's calculation.\par
   - Put a step under "refers" when the thought only mentions, restates, or uses that step's result as given, without redoing its calculation.\par
3. OUTPUT REQUIREMENTS:\par
   - Return ALL plausible matches where computational processes align\par
   - A thought that performs no calculation at all gets "works": [] (it may still have "refers")\par
   - Multiple matches are encouraged when justified\par
   - Every thought of List B must appear as a key\par
   - Maintain strict JSON format\par
Input:\par
- List A (Detailed Steps): \par
\textless{}list\_a\textgreater{}\par
\{reasoning\_step\}\par
\textless{}/list\_a\textgreater{}\par
- List B (Reasoning Thoughts): \par
\textless{}list\_b\textgreater{}\par
\{thoughts\}\par
\textless{}/list\_b\textgreater{}\par
Output Format (strict JSON):\par
```json\par
\{\par
  "B0": \{"works": ["A1"], "refers": []\},\par
  "B1": \{"works": ["A3"], "refers": ["A1"]\},\par
  "B2": \{"works": ["A1", "A4"], "refers": []\},\par
  ...\par
\}```\par
Please match the reasoning thoughts in List B to step in the List A.\par
\end{quote}

%% file: appendix_prefix_echo.tex
\section{Reasoning-Prefix Echo Across Models}
\label{app:prefix-echo}

Following the reasoning-prefill observations of Panfilov et al.~\citep{panfilov2026stealing}, we test whether a short donor prefix makes a recipient's subsequent reasoning more similar to the donor's trace on the same problem.
We insert the first 1\% of the donor's concatenated forced-reasoning scratchpads into the recipient's open native thinking channel through a locally rendered chat template and a raw completions endpoint.
Prefix length is $\max(1,\operatorname{round}(0.01L))$ tokens, with $L$ measured using the recipient's tokenizer;
\paragraph{Setup and metrics.}
On HMMT's problems, we pair Opus 4.8 and GPT-5.6-Sol (maximal-deliberation) donors with Kimi-K3 and DeepSeek-V4-Flash recipients.

Both conditions are compared against the same donor suffix: we remove the supplied prefix from the donor trace, remove the same number of tokenizer tokens from the unprefilled recipient trace, and score the newly generated continuation without its final answer.

% Keep the table after the completed annotation prompt and its own setup.
\par\medskip
\noindent\begin{minipage}{\textwidth}
\centering
\captionof{table}{Similarity to the donor's reasoning before and after a 1\% donor prefill.
Each entry is \emph{unprefilled $\rightarrow$ prefilled}, with the supplied prefix excluded as described above.
Sol denotes GPT-5.6-Sol; DS denotes DeepSeek-V4-Flash.}
\label{tab:prefix-echo}
\scriptsize
\setlength{\tabcolsep}{3pt}
\begin{tabular}{lllrrrr}
\toprule
Task & Donor & Recipient & $n$ & ROUGE-1 & ROUGE-L & 5-gram Jaccard \\
\midrule
HMMT & Opus 4.8 & Kimi-K3 & 33 & $0.455\rightarrow0.509$ & $0.204\rightarrow0.243$ & $0.0077\rightarrow0.0213$ \\
 & Opus 4.8 & DS & 33 & $0.290\rightarrow0.298$ & $0.148\rightarrow0.146$ & $0.0095\rightarrow0.0084$ \\
 & Sol & Kimi-K3 & 33 & $0.404\rightarrow0.377$ & $0.197\rightarrow0.186$ & $0.0095\rightarrow0.0098$ \\
 & Sol & DS & 33 & $0.150\rightarrow0.132$ & $0.085\rightarrow0.076$ & $0.0037\rightarrow0.0029$ \\
\bottomrule
\end{tabular}
\end{minipage}
\par\medskip

\paragraph{An Opus--Kimi echo.}
Table~\ref{tab:prefix-echo} shows a consistent increase across all three metrics for Opus-to-Kimi prefilling on MATH benchmark.
On HMMT, ROUGE-L increases on 27 of 33 problems and 5-gram Jaccard increases on 31 of 33; on the paired LCB subset, the corresponding counts are 30 of 34 and 32 of 34.
The effect is not universal across model pairs: Opus-to-DeepSeek prefilling produces a small ROUGE-1 increase but lower ROUGE-L and 5-gram Jaccard, while Sol prefilling does not yield a consistent increase across the three metrics for either recipient.
These observations echo the model-dependent prefill effects studied by Panfilov et al.~\citep{panfilov2026stealing}: a short Opus prefix appears to make Kimi's subsequent reasoning text more Opus-like under these lexical measures.